\pdfoutput=1

\documentclass[11pt]{article}

\usepackage[]{acl}

\usepackage{times}
\usepackage{latexsym}
\usepackage[T1]{fontenc}

\usepackage[utf8]{inputenc}
\usepackage{microtype}
\usepackage{inconsolata}
\usepackage{enumitem}
\usepackage[disable]{todonotes}
\usepackage{hyperref}
\usepackage{multirow}
\usepackage{multicol}
\usepackage{booktabs}
\usepackage{lipsum}
\usepackage{dsfont}
\usepackage{amsthm}
\usepackage{amsmath}
\usepackage{algorithm}
\usepackage{algorithmic}
\usepackage{graphicx}
\usepackage{caption}
\usepackage{subcaption}
\usepackage{diagbox}
\usepackage{slashbox}
\hypersetup{
  colorlinks   = true, 
  urlcolor     = purple, 
  linkcolor    = magenta, 
  citecolor   = purple 
}
\newtheorem*{remark}{Remark}
\newcommand{\model}{\textsc{SMART}}
\newtheorem{definition}{Definition}

\title{\model{}: Submodular Data Mixture Strategy for Instruction Tuning}

\author{
    H S V N S Kowndinya Renduchintala \\
    Media and Data Science Research\\
    Adobe Inc., India\\
    \texttt{rharisrikowndinya333@gmail.com} \\\AND
    Sumit Bhatia \\
    Media and Data Science Research\\
    Adobe Inc., India\\
    \texttt{sumit.bhatia@adobe.com} \\\And
    Ganesh Ramakrishnan \\
    Dept. of Computer Science \& Engineering\\
    Indian Institute of Technology Bombay\\
    \texttt{ganesh@cse.iitb.ac.in} \\
}
  
\begin{document}
\maketitle
\begin{abstract}
\textit{Instruction Tuning} involves finetuning a language model on a collection of instruction-formatted datasets in order to  enhance the generalizability of the model to unseen tasks. Studies have shown the importance of balancing different task proportions during finetuning, but finding the right balance remains challenging. Unfortunately, there's currently no systematic method beyond manual tuning or relying on practitioners' intuition. In this paper, we introduce \model{} (\textbf{S}ubmodular data \textbf{M}ixture str\textbf{A}tegy for inst\textbf{R}uction \textbf{T}uning) --- a novel data mixture strategy which makes use of a submodular function to assign importance scores to tasks which are then used to determine the mixture weights. Given a fine-tuning budget, \model{} redistributes the budget among tasks and selects non-redundant samples from each task. Experimental results demonstrate that \model{} significantly outperforms traditional methods such as examples proportional mixing and equal mixing. Furthermore, \model{} facilitates the creation of data mixtures based on a few representative subsets of tasks alone and through task pruning analysis, we reveal that in a limited budget setting, allocating budget among a subset of representative tasks yields superior performance compared to distributing the budget among all tasks. The code for reproducing our results is open-sourced at \href{https://github.com/kowndinya-renduchintala/SMART}{https://github.com/kowndinya-renduchintala/SMART}.
\end{abstract}

\section{Introduction}
\label{sec:introduction}
\textit{``Your ability to juggle many tasks will take you far.''}

\vspace{\baselineskip}
One of the main goals of artificial intelligence (AI) research is to build machines that can \textit{communicate} \citep{turing_computing_1950}, and an essential part of communication is to understand and follow \textit{instructions}. Large Language Models (LLMs), which are pre-trained over massive text corpora on \textit{next-token-prediction} objective, can perform a wide range of NLP tasks via \textit{``prompting''}~\citep{brown2020language,kojima2022large,almazrouei2023falcon,liu2023pre,touvron2023llama}.

\textit{Instruction Tuning}~\citep{wei2021finetuned,sanh2021multitask,chung2022scaling} is an approach that further enhances the instruction-following ability and generalizability of pre-trained LLMs to unseen tasks. It involves fine-tuning an LLM on a collection of instruction-formatted instances (encompassing multiple tasks) - each consisting of an instruction (or task description), an optional input, the corresponding output (the ground truth) and optionally a few demonstrations/examples. It is a special case of multitask learning where the LLM is finetuned on a collection of instruction-formatted multitask datasets~\cite{chung2022scaling}. Finetuning on multiple tasks simultaneously, allows the model to share and transfer information across tasks, resulting in a better common internal representation that is preferred by all tasks while suppressing task-dependent noise~\citep{caruana1997multitask}. Consequently, the model learns to generalize to unseen tasks by discerning helpful cues from both implicitly and explicitly related tasks that it has previously seen.

The performance enhancement from instruction tuning is heavily contingent on data quality, data quantity, and task composition~\citep{wang2023data}. Studies by \citet{iyer2022opt} and \citet{longpre2023flan} have shown that while scaling the number of tasks is important, the relative proportion of various tasks (mixture weighting) merits as much attention for optimal instruction tuning. Intuitively, we want the model to see enough data for a given task that it can perform well on it, but not to see so much data that it memorizes the training set~\citep{raffel2020exploring}. ~\citet{iyer2022opt} performed \textit{manual tuning} of various benchmark proportions and decided on a final mixture, whereas \citet{longpre2023flan} studied the impact of removing each benchmark from the finetuning mixture and relied on their \textit{practioners' intuition} from there on, to decide on the exact proportions of benchmarks. In this work, we would like to explore a more systematic approach to mixture weighting. Specifically, we are motivated by the fact that in a large multitask dataset like FLAN 2022~\citep{longpre2023flan}, which has 1840 tasks, there will likely be many similar tasks leading to redundancies and not all of them may require sampling in equal proportions. For instance, there might be many tasks of the type Natural Language Inference (NLI), and it might be enough to sample relatively more instances from a few \textit{representative} NLI tasks and less from the others. Furthermore, \textit{which} samples we select from each task is also crucial because the samples should faithfully represent the task at hand. A random subset may fail to do this as it can miss out on essential corner cases.

With this context, we focus on the following two fundamental research questions (\textbf{RQ}s) that form the basis for our subsequent inquiry:
\begin{itemize}[nosep]
    \item \textbf{(RQ1)} Given a \textit{huge} multitask instruction-tuning dataset and a limited fine-tuning budget which is defined by the total number of $(prompt, response)$ instances that can be used for fine-tuning, how do we divide this budget among thousands of tasks present in the dataset? i.e., \textit{how many} instances to sample from each task? and \textit{which} instances to sample from each task?
    \item \textbf{(RQ2)} Can we go a step further and strategically prune some tasks altogether and only fine-tune on a small subset of representative tasks without hurting the performance? If yes, what is the nature of this subset?
\end{itemize}

To the best of our knowledge, there's currently no principled approach to determining task compositions for instruction tuning, other than \textit{manual tuning} and/or \textit{practioners' intuition}.

As a first step towards addressing both of the above \textbf{RQ}s, we first define a common subset selection problem (more formally stated in Section~\ref{sec:approach}) as follows - Given a \textit{huge} collection of $M$ instruction-formatted task datasets, a task budget $M' \le M$ and a total budget ($N'$) of $(prompt, response)$ pairs, which $M'$ tasks to select? and how many instances to select from each of these $M'$ tasks and which instances to select? Note that \textbf{RQ1} is an instance of this problem where $M'=M$.

Constrained Submodular Maximization (Section~\ref{sec:background_submodularity}) proves to be a good model for discovering representative subsets (or \textit{coresets}) of a massive training dataset (or \textit{ground set)} that acts as surrogate (i.e., achieves similar performance) and are much better than uniformly-at-random subsets. Intuitively, this is because submodular functions model \textit{information} in subsets, and hence maximizing a submodular function subject to a constraint yields non-redundant subsets of the ground set. An essential feature of this model is that it returns weighted subsets, i.e., each sample in the coreset comes with an associated score, which indicates how important the sample is. 

Inspired by submodular functions, we propose our solution (Section~\ref{sec:approach}) to the above subset selection problem for instruction tuning that works in two stages. In the first stage, we select a weighted subset of tasks from the full dataset where the weights will determine how many samples to select from each task. In the next stage, we select samples from each task based on the assigned task budgets. Note that the submodular functions used in each stage are not necessarily identical (Section~\ref{sec:ablation_submodular_function}). 

The \textit{main contributions} of our work can be summarized as follows:
\begin{itemize}[nosep]
    \item We introduce \model{} --- a novel data mixture strategy for instruction tuning that models the data mixture problem (Section~\ref{sec:approach}) as a sequence of two cardinality-constrained submodular maximization problems and offer empirical evidence that it outperforms both examples proportional and equal mixing baselines (Section~\ref{sec:experiments}) as well as the mixture weights proposed by \citet{longpre2023flan}.
    \item Existing works like \citet{longpre2023flan} have reported a continuous increase in performance upon increasing the number of tasks (though the gains themselves may be diminishing). However, we posit that this depends on the order in which new tasks are incorporated and show empirically that in the case of \model{} mixtures, a performance peak is observed with an initial addition of few representative tasks and upon adding more and more tasks, the performance is not sustained (Section ~\ref{sec:addressing_rq2}).
    \item We find that the nature of instances that should be selected in each task (i.e, whether a representative or diverse subset) also depends on the total task budget, $M'$ (Section~\ref{sec:ablation_submodular_function}). For higher $M'$s, each task on average gets the relatively low budget and selecting representative samples is more important; however for lower $M'$s, when there is sufficient enough budget for each task, the need for diversity dominates that of representation.
\end{itemize}

\newpage
\section{Background: Submodularity}
\label{sec:background_submodularity}
\paragraph{Notations.}
Let $f:2^{\mathcal{V}}\to\mathds{R}$ be a \textit{set function} that assigns a value to every subset of the \textit{ground set} $\mathcal{V}$. We use the notation $f(v|X)$ as a shorthand for $f(X \cup \{v\})-f(X)$ i.e., the incremental value \textit{gain} of $v$ in the context of $X$. 

\begin{definition}
    (\textit{Submodular Function}) A given set function $f:2^{\mathcal{V}}\to\mathds{R}$ is submodular if for all $X, Y \subseteq \mathcal{V}$, where $X \subseteq Y$ and for all $v \notin Y$, the following inequality holds true:
    \[
        f(v|X) \ge f(v|Y) \tag{diminishing gains property}
    \]
\end{definition}

Intuitively, the definition states that --- Adding an element to a smaller set $X$ yields more value gain than adding the same element to a superset $Y$ of $X$. Table~\ref{tab:submodular_functions_examples} contains examples of three submodular functions --- Facility Location (models representation), Log Determinant (models diversity), and Graph Cut (models a trade-off between representation and diversity controlled by the parameter $\lambda$). 

\begin{table}[h!]
    \centering
    \begin{tabular}{|c|c|}
        \hline
        Submodular Function & $f(X)$ \\
        \hline
        Facility Location & $\underset{i\in\mathcal{V}}{\sum}{\underset{j \in X}{\max}{ s_{ij}}}$ \\
        Graph Cut & $\underset{i \in \mathcal{V}, j \in X}{\sum}{s_{ij}} - \lambda\underset{i, j \in X}{\sum}{s_{ij}}$ \\
        Log Determinant & $\log\det(\mathcal{S}_X)$ \\
        \hline
    \end{tabular}
    \caption{Examples of Submodular Functions. $\mathcal{V}$ is the ground set and $X\subseteq\mathcal{V}$. $s_{ij}$ is the similarity between two elements $i$ and $j$ of the ground set. $\mathcal{S}_X$ is the similarity matrix between the items in $X$. Facility Location models representation; Log Determinant models diversity and Graph Cut models a trade-off between  representation and diversity (governed by the parameter $\lambda$).}
    \label{tab:submodular_functions_examples}
\end{table}
\begin{definition}
    (\textit{Cardinality Constrained Submodular Maximization}) Given a submodular function $f:2^{\mathcal{V}}\to\mathds{R}$ defined over the subsets of the ground set $\mathcal{V}$, the constrained submodular maximization problem involves finding $\mathcal{S}^{*}$ such that
    \[
        \mathcal{S}^{*} = \underset{\substack{X\subseteq\mathcal{V} \\ |X|\le N'}}{\arg \max}{f(X)}
    \]
\end{definition}

The above cardinality-constrained submodular maximization problem is NP-complete~\cite {feige1998threshold}. However, if $f$ is monotone submodular (i.e., $f(X)\le f(Y)$ whenever $X\subseteq Y$), \citealp{nemhauser1978analysis,fisher1978analysis} show that a simple greedy algorithm described in Algorithm~\ref{alg:greedy_algorithm} can be used to find an approximate solution $\mathcal{S}^{greedy}$, with a guarantee that $f(\mathcal{S}^{greedy}) \ge (1-1/e)f(\mathcal{S}^{*})$.\footnote{Assuming $P\neq NP$, this is the best approximation ratio that can be achieved by any polynomial time algorithm.}

\begin{algorithm}
    \caption{The Na\"ive Greedy}
    \label{alg:greedy_algorithm}
    \begin{algorithmic}
        \STATE {\bfseries Input:} Ground Set ($\mathcal{V}$), Budget ($N'$)
        \STATE $X_0 \gets \emptyset$;
        \STATE $\mathcal{S} \gets [\,]$;
        \STATE $Gains \gets [\,]$;
        \FOR{$i=0$ {\bfseries to} ($N'-1$)}
            \STATE $e^* = \underset{v\in \mathcal{V}\setminus X_{i}}{\arg \max}{f(v|X_i})$;
            \STATE $g_{i+1} = f(e^* | X_{i})$;
            \STATE $X_{i+1} = X_{i} \cup \{e^*\}$;
            \STATE $\mathcal{S}\text{.append}(e^*)$;
            \STATE $Gains\text{.append}(g_{i+1})$;
        \ENDFOR
        \RETURN $\mathcal{S}, Gains$;
    \end{algorithmic}
\end{algorithm}

\begin{remark}
The algorithm produces a weighted subset where the value gains themselves are the weights (i.e., $\underset{v\in \mathcal{V}\setminus X_{i}}{\max}{f(v|X_i})$ is the weight of $\underset{v\in \mathcal{V}\setminus X_{i}}{\arg \max}{f(v|X_i})$). 
\end{remark}

Algorithm~\ref{alg:greedy_algorithm} (also known as Na\"ive Greedy) requires $\mathcal{O}(N'.|\mathcal{V}|)$ function evaluations which is costly in practice. Accelerated Greedy~\citep{minoux2005accelerated}, also known as Lazy Greedy, can be used instead, which leverages submodularity and offers a more efficient heap-based implementation of the same algorithm.

\section{Approach}
\label{sec:approach}
Inspired by submodularity (Section ~\ref{sec:background_submodularity}), in this section, we introduce a novel data mixture strategy, \model{},  as a technique to solve the following subset selection problem introduced in Section ~\ref{sec:introduction}: 

\textit{Consider a collection $\mathcal{D}=\{\mathcal{T}_1, \dots, \mathcal{T}_M\}$ of $M$ instruction-formatted task datasets where each $\mathcal{T}_i=\{(prompt_{ij}, response_{ij})\}_{j=1}^{N_{\mathcal{T}_i}}$ consists of $N_{\mathcal{T}_i}$ $(prompt, response)$ pairs. Let $\sum_{i=1}^{M}N_{T_i} = N$. Given an $M'\le M$ and an $N'\le N$, how do we select a subset of tasks $\mathcal{D}'=\{\mathcal{T}_{1}', \dots, \mathcal{T}_{M'}'\}$(where $\mathcal{D}' \subseteq \mathcal{D}$), and subsequently $\mathcal{S}=\{\mathcal{S}_1, \dots, \mathcal{S}_{M'}\}$ (where $\mathcal{S}_j\subseteq\mathcal{T}_{j}'$) and $\sum_{j=1}^{M'}|\mathcal{S}_j|=N'$) such that efficiently fine-tuning on the subset $\mathcal{S}$ alone is (nearly) as effective as fine-tuning on the entire collection $\mathcal{D}$?} 

\model{} models the above problem as a sequence of two cardinality-constrained submodular maximization problems. The first one is to select a weighted subset of $M'$ tasks and the second one is to select instances - a total of $N'$ instances from these tasks. The weights obtained in the first stage will be used to determine how many instances we sample from each task.

\subsection{The Algorithm}

We now give a detailed description of the two stages of \model{} (summarized in Algorithm ~\ref{alg:main_algorithm}):
\subsubsection{Stage-1: Weighted Task Subset Selection}
\label{sec:weighted_task_subset_selection}
In this first stage, given the instruction-tuning dataset $\mathcal{D}=\{\mathcal{T}_1, \dots, \mathcal{T}_M\}$, our goal is to find $\mathcal{D}'=\{\mathcal{T}_{1}', \dots, \mathcal{T}_{M'}'\}$ where $\mathcal{D}' \subseteq \mathcal{D}$, along with the instance budgets, $\{N_{1}', \dots, N_{M'}'\}$, such that $\sum_{j=1}^{M'}|N_{j}'|=N'$. 

If $f_1$ is the submodular function that we use in this stage, $\mathcal{D}'$ is given by:
\[
    \mathcal{D}'=\underset{\substack{X\subseteq \mathcal{D} \\ |X|\le M'}}{\arg \max}{f_1(X)}
\]
To find the instance budgets ($N_{j}'$s), we use the second-order Taylor-softmax operation \citep{de2015exploration} on the value gains obtained from the greedy algorithm, to compute a probability distribution which determines the probability with which instances will be sampled from a given task i.e., if $\{g_1, \dots g_{M'}\}$ are the value gains returned by the greedy algorithm, corresponding to the tasks $\{\mathcal{T}_{1}', \dots, \mathcal{T}_{M'}'\}$, the instance budgets are given by
\[
    N_{j}'=\frac{(1+g_j+0.5g_j^2)}{\sum_{k=1}^{M'}(1+g_k+0.5g_k^2)} \times N'
\]

\subsubsection{Stage-2: Instance Subset Selection}
\label{sec:instance_subset_selection}
In this stage, given the subset of tasks, $\{\mathcal{T}_{1}', \dots, \mathcal{T}_{M'}'\}$, and the instance budgets $\{N_{1}', \dots N_{M'}'\}$ from the first stage, the goal is to actually select those many samples from each task. If $f_2$ is the submodular function used, the final subset $\mathcal{S}$ is given by
\[
    \mathcal{S}=\bigcup_{j=1}^{M'}\underset{\substack{X_j\subseteq \mathcal{T}_{j}' \\ |X_j|\le N_{j}^{'}}}{\arg \max}{f_2(X_j)}
\]

\begin{algorithm}[h]
    \caption{The \model{} Data Mixture Strategy}
    \label{alg:main_algorithm}
    
    \begin{algorithmic}
        \STATE {\bfseries Input:} Task datasets $\mathcal{D}=\{\mathcal{T}_1, \dots, \mathcal{T}_M\}$, Task Budget ($M'$), Instance Budget ($N'$), $f_1$, $f_2$ \;
        
        \textcolor{gray}{\# Each $\mathcal{T}_i=\{(prompt_{ij}, response_{ij})\}_{j=1}^{N_{\mathcal{T}_i}}$ } \;
        \STATE $encoder \gets \text{SentenceEncoder()}$ \;
        \STATE $prompt\_embeddings \gets \text{dict()}$ \;
        \STATE $task\_embeddings \gets \text{dict()}$ \;
        \FOR{$i=1$ {\bfseries to} $M$}
            \STATE $\mathbf{V} \gets [\,]$ \;
            \FOR{$j=1$ {\bfseries to} $N_{\mathcal{T}_i}$}
                \STATE $\mathbf{v} \gets encoder\text{.encode}(prompt_{ij})$ \;
                \STATE $\mathbf{V}\text{.append}(\mathbf{v})$ \;
            \ENDFOR
            \STATE $prompt\_embeddings[\mathcal{T}_{i}]=\mathbf{V}$ \;
            \STATE $task\_embeddings[\mathcal{T}_i]=\dfrac{1}{N_{\mathcal{T}_i}}\sum\limits_{j=0}^{N_{\mathcal{T}_i}-1}\mathbf{V}[j]$ \;
        \ENDFOR
        \STATE $\mathcal{K}_{task}=\text{cos\_sim}(task\_embeddings)$ \;
        \STATE $\mathcal{D}', gains = \text{Greedy}(f_1, \mathcal{D}, \mathcal{K}_{task}, M')$ \;
    
        \textcolor{gray}{\# $\mathcal{D}'=\{\mathcal{T}_{1}', \dots \mathcal{T}_{M'}'\}$} \;

        \STATE $probs \gets \text{Taylor\_Softmax}(gains)$ \;
        
        \STATE $\mathcal{S}'\gets [\,]$ \;
        \FOR{$j=1$ {\bfseries to} $M'$}
            \STATE $N_j'=probs[j]\times N'$
            \STATE $\mathcal{K}_{\mathcal{T}_j'}=\text{cos\_sim}(prompt\_embeddings[\mathcal{T}_j'])$ \;
            \STATE $\mathcal{S}_{\mathcal{T}_j'}, gains = \text{Greedy}(f_2, \mathcal{T}_j', \mathcal{K}_{\mathcal{T}_j'}, N_{j}')$ \;
            \STATE $\mathcal{S}'\text{.append}(\mathcal{S}_{\mathcal{T}_j'}) $
        \ENDFOR
        \STATE $\mathcal{S}=\bigcup\limits_{j=1}^{M'}\mathcal{S}'[j]$
        \RETURN $\mathcal{S}$
    \end{algorithmic}
\end{algorithm}
\subsection{Obtaining Similarity Measures}
\label{sec:obtaining_sij}

The three submodular functions listed in Table~\ref{tab:submodular_functions_examples} require computation of similarity measures ($s_{ij}$) between items in the ground set. So, all prompts in the dataset are first encoded using a sentence encoder. For instance subset selection (i.e., \textbf{Stage-2}), cosine similarity between prompt embeddings is used as the similarity measure and for weighted task subset selection (i.e., \textbf{Stage-1}), cosine similarity between task embeddings is computed, where the task embeddings are computed as the average prompt embeddings, following \citet{vu2020exploring}. Although there are other sophisticated methods ~\citep{achille2019task2vec,zhou2022efficiently,xi2023connectivity,vu2021spot} for obtaining task embeddings, most of them depend on the LLM at hand.

\subsection{Choosing $f_1$ and $f_2$} 
\label{sec:choosing_f1_f2}

Each submodular function in Table~\ref{tab:submodular_functions_examples} captures a different property: facility-location emphasizes representation, graph-cut balances representation and diversity, and log-determinant prioritizes diversity. We treat $f_1$ and $f_2$ as hyperparameters in our grid search in Section~\ref{sec:ablation_submodular_function}, exploring three options for each function. Determining the optimal $f_1$ and $f_2$ is part of our research question (\textbf{RQ2}). We discuss the findings of grid search and their qualitative implications for instruction tuning in Section~\ref{sec:insights_and_discussions}.

\section{Experiments}
\label{sec:experiments}
\subsection{Finetuning Data}
\label{sec:finetuning_data}
In all the experiments, the underlying ground set ($\mathcal{D}$) is FLAN 2022~\citep{longpre2023flan,chung2022scaling}. The collection consists of the following five sub mixtures adding up to a total of 1840 tasks and 17,591,640 ($\sim$17.5M) instruction-formatted $(prompt, response)$ pairs:
\begin{itemize}[nosep]
    \item FLAN 2021~\citep{wei2021finetuned}
    \item T0 ~\citep{sanh2021multitask}
    \item NIV2~\citep{wang2022super}
    \item CoT (several chain-of-thought datasets)
    \item Dialog (a few dialog datasets)
\end{itemize}
Each of the tasks comes in a variety of templates - zeroshot with and without answer options, fewshot with and without answer options.

\paragraph{\model{} Data Mixture Creation} We encode prompts in the collection with GTE-large~\citep{li2023towards}, a light-weight (340M parameters) BERT-based effective ~\citep{muennighoff2022mteb} sentence encoder for semantic textual similarity. Task embeddings are obtained by averaging the corresponding prompt embeddings (Section~\ref{sec:obtaining_sij}). We use SUBMODLIB~\citep{kaushal2022submodlib}, which has the necessary algorithms implemented, to obtain the weighted task subsets (Section ~\ref{sec:weighted_task_subset_selection}) and the instances (Section ~\ref{sec:instance_subset_selection}). We distribute the instance budgets equally among task templates based on findings by \citet{longpre2023flan} which showed that an equal number of zero-shot and few-shot templates yield the best performance on held-out tasks.

\subsection{Finetuning Procedure}
\label{sec:finetuning_procedure}
We evaluate \model{} on three 7B parameter LLMs: Llama-2~\cite{touvron2023llama}, Falcon~\cite{almazrouei2023falcon}, and Mistral~\cite{jiang2023mistral}. We fine-tune each model for 1 epoch on a given data mixture with a learning rate of $2e-5$ for Llama-2 and Falcon, and $5e-6$ for Mistral. A batch size of $64$, weight decay of $0.1$, and cosine learning rate decay with a linear warmup for the initial $1\%$ training steps are employed. The experiments all ran on 8 NVIDIA A100-SXM4-80GB GPUs, utilizing Flash Attention~\cite{dao2023flashattention2} for memory efficiency and speeding up finetuning. Our code is open-sourced \href{https://github.com/kowndinya-renduchintala/SMART}{here}.

\subsection{Baselines}
\label{sec:baselines}

We mainly compare \model{} with two baseline mixture strategies: Examples Proportional Mixture, Equal Mixture ~\citep{raffel2020exploring}.

\paragraph{Examples Proportional Mixture (Baseline-1)} Instances are sampled in proportion to the size of each task's dataset. This is equivalent to randomly sampling from the combined datasets.
\paragraph{Equal Mixture (Baseline-2)} Instances are sampled from each task with equal probability {\em i.e.}, by dividing the total budget equally among tasks, and then uniformly sampling from each task.

We also compare \model{} with mixture weights used by \citet{longpre2023flan} but since their mixture weights apply to the five sub mixtures listed in Section~\ref{sec:finetuning_data} rather than individual tasks, we analyse this separately in Section~\ref{sec:comparison_with_flan_v2_mix}.

\begin{table*}[hbt!]
    \centering
    \resizebox{\textwidth}{!}{
        \begin{tabular}{lcccccccccc}
            \toprule
            \multirow{4}{*}{$N'$} & \multirow{4}{*}{Data Mix.} & \multicolumn{5}{c}{MMLU-ZeroShot (Exact Match)} & \multicolumn{3}{c}{BBH-Zeroshot (Exact Match)} & MMLU + BBH \\
            \cmidrule(lr){3-7}\cmidrule(lr){8-10}\cmidrule{11-11}
            & & STEM & Humanities & \begin{tabular}{c} Social \\ Sciences\end{tabular} & Other & \begin{tabular}{c}MMLU \\ FULL\end{tabular} & NLP & Algorithmic & \begin{tabular}{c}BBH \\ FULL\end{tabular} & (Weighted Avg.) \\
            \midrule
            \multirow{3}{*}{$25000$} & EPM (Baseline-1) & 30.82 & 46 & 46.71 & 43.05 & 40.63 & \textbf{40.59} & \textbf{25.55} & \textbf{31.67} & 38.05 \\
             & EM (Baseline-2) & 30.33 & 45.01 & 43.81 & 40.55 & 39.03 & 40.08 & 20.29 & 29.46 & 36.27 \\
             % & FLANv2 (Baseline-3) & 30.55 & 49.08 & 50.14 & 45.09 & 42.47 & \textbf{43.79} & \textbf{25.73} & \textbf{33.33} & \textbf{39.84} \\
             & \model{} (Ours) & \textbf{32.22} & \textbf{50.41} & \textbf{50.14} & \textbf{46.85} & \textbf{43.73} & 38.85 & 24.16 & 30.05 & \textbf{39.8} \\
            \hline
            \multirow{3}{*}{$50000$} & EPM (Baseline-1) & 31.59 & 47.68 & 47.18 & 44.68 & 41.76 & 41.25 & \textbf{26.49} & \textbf{32.64} & 39.14 \\
             & EM (Baseline-2) & 35.22 & 49.58 & 51.01 & 48.13 & 44.99 & 41.96 & 22.83 & 31.24 & 41.04 \\
             % & FLANv2 (Baseline-3) & 31.85 & 52.54 & 51.71 & 47.54 & 44.6 & 43.13 & 26.13 & 33.01 & 41.27 \\
             & \model{} (Ours) & \textbf{36.51} & \textbf{53.06} & \textbf{54.49} & \textbf{50.79} & \textbf{47.58} & \textbf{46.73} & 20.1 & 31.75 & \textbf{43.03} \\
            \hline
            \multirow{3}{*}{$100000$} & EPM (Baseline-1) & 32.66 & 50.6 & 51.37 & 47.18 & 44.25 & 43.38 & \textbf{26.36} & 33.48 & 41.16\\
             & EM (Baseline-2) & 36.03 & 50.7 & 52.53 & 47.1 & 45.57 & 44.4 & 25.18 & 33.55 & 42.11\\
             % & FLANv2 (Baseline-3) & 36.1 & 54.1 & \textbf{55.6} & 50.44 & 47.82 & 44.44 & 26.22 & 33.42 & 43.68\\
             & \model{} (Ours) & \textbf{37.36} & \textbf{55.38} &\textbf{55.47} & \textbf{52.95} & \textbf{49.11} & \textbf{47.26} & 24.22 & \textbf{34.66} & \textbf{44.96} \\
            \hline
            \multirow{3}{*}{$200000$} & EPM (Baseline-1) & 35.19 & 54.64 & 54.75 & 50.58 & 47.53 & 45.25 & \textbf{26.57} & 34.52 & 43.79 \\
             & EM (Baseline-2) & 38.6 & 54.05 & 54.72 & 51.47 & 48.68 & 41.36 & 24.75 & 32.28 & 43.96 \\
             % & FLANv2 (Baseline-3) & 34.76 & 55.02 & 56.65 & 50.13 & 47.76 & 44.39 & \textbf{27.68} & 34.73 & 44.02 \\
             & \model{} (Ours) & \textbf{39.2} & \textbf{57.29} & \textbf{58.71} & \textbf{55.01} & \textbf{51.32} & \textbf{47.99} & 24.47 & \textbf{35.27} & \textbf{46.7} \\
            \hline
            \multirow{3}{*}{$400000$} & EPM (Baseline-1) & 38.16 & 56.53 & 56.99 & 52.56 & 49.85 & 48.72 & 26.04 & 36.49 & 46.01 \\
             & EM (Baseline-2) & 39.43 & 55.97 & 57.59 & 53.65 & 50.52 & 47.37 & 26.08 & 35.8 & 46.29 \\
             % & FLANv2 (Baseline-3) & 37.46 & 56.79 & 58.42 & 51.4 & 49.71 & 49.11 & \textbf{26.56} & 36.97 & 46.05 \\
             & \model{} (Ours) & \textbf{39.77} & \textbf{57.39} & \textbf{60.17} & \textbf{54.79} & \textbf{51.77} & \textbf{49.25} & \textbf{26.35} & \textbf{37.43} & \textbf{47.65} \\
            \hline
            17,591,640 & Full FLAN 2022 & 42.44 & 59.1 & 61.82 & 55.1 & 53.43 & 50.7 & 27.6 & 38.11 & 49.03 \\
            \hline
        \end{tabular}
    }
    \caption{Comparison of \model{} with baselines on MMLU-zeroshot and BBH-zeroshot for Llama-2-7b. EPM (Baseline-1) denotes Examples Proportional Mixture and EM (Baseline-2) denotes Equal Mixture. All the scores are Exact Matches. Weighted average on 57 MMLU tasks and 23 BBH tasks is reported in the last column. For baselines, exact matches are obtained by averaging across 3 fine-tuning runs.}
    \label{tab:main_table_rq1}
\end{table*}

\subsection{Evaluation Protocol}
\label{sec:evaluation_protocol}

We evaluate the fine-tuned models on two benchmark datasets: MMLU~\cite{hendrycks2020measuring} with 57 tasks assessing world knowledge and problem-solving, and BBH~\cite{suzgun2022challenging} with 23 challenging tasks from Big-Bench~\cite{srivastava2022beyond}. MMLU covers STEM, Humanities, Social Sciences, and Other (business, medical, and misc.) categories, while BBH includes both NLP and Algorithmic tasks\footnote{The \textit{Algorithmic} subcategory is so named because these tasks (e.g., 2-digit arithmetic)  do not require an LLM to be solved.}. Evaluation involves prompting the LLM directly and using Exact Match as the scoring metric. Responses are generated using the greedy decoding approach and they undergo basic post-processing steps (removing punctuation and lower-casing) before calculating exact match. For baseline data mixtures, 3 mixtures with different random seeds are created and the mean exact match of the 3 fine-tuning runs is reported.

\subsection{Addressing \textbf{RQ1} ($M'=M$)}
\label{sec:addressing_rq1}

\textbf{RQ1} is an instance of the subset selection problem defined in Section ~\ref{sec:approach}, where $M'=M$, allowing all tasks for finetuning; however we still have a constraint ($N'$) on total number of $(prompt, response)$ pairs. We use Graph-Cut and Facility Location functions (Table~\ref{tab:submodular_functions_examples})  in Stage-1 (Section~\ref{sec:weighted_task_subset_selection}) and Stage-2 (Section~\ref{sec:instance_subset_selection}) of \model{} respectively. This choice of $f_1$ and $f_2$ is determined via grid search discussed in Section~\ref{sec:ablation_submodular_function}.
Table~\ref{tab:main_table_rq1} contains comparison of \model{} mixtures with baseline mixtures, on MMLU and BBH benchmarks, upon instruction fine-tuning Llama-2-7b on data mixtures generated by varying $N'$ in $\{25000, 50000, 100000, 200000, 400000\}$. \model{} data mixtures consistently perform better than both examples proportional mixtures and equal mixtures baseline.

\begin{figure*}[hbt!]
    \centering
    \begin{subfigure}[b]{0.32\textwidth}
        \centering
        \includegraphics[width=\linewidth]{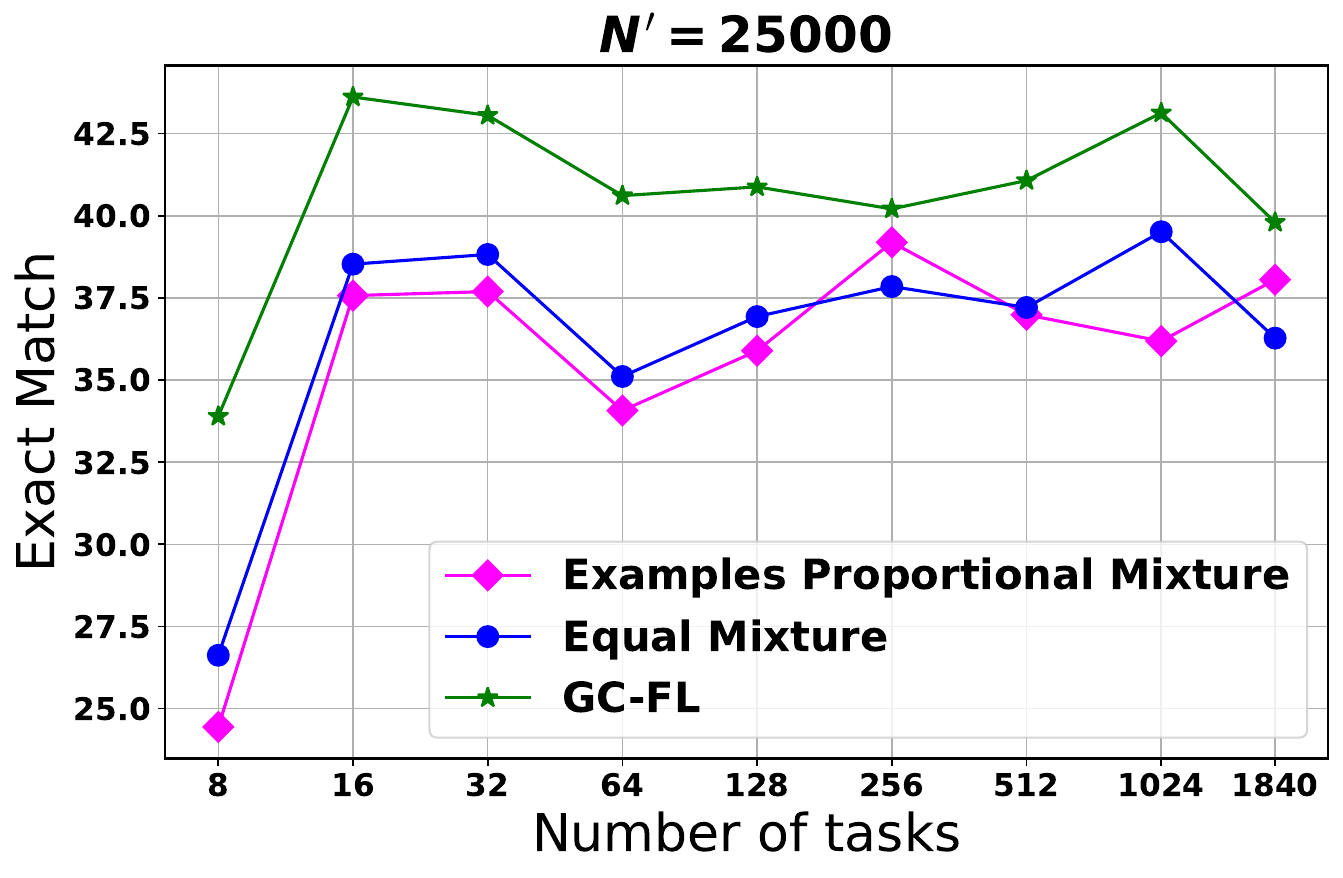}
    \end{subfigure}
    \begin{subfigure}[b]{0.32\textwidth}
        \includegraphics[width=\linewidth]{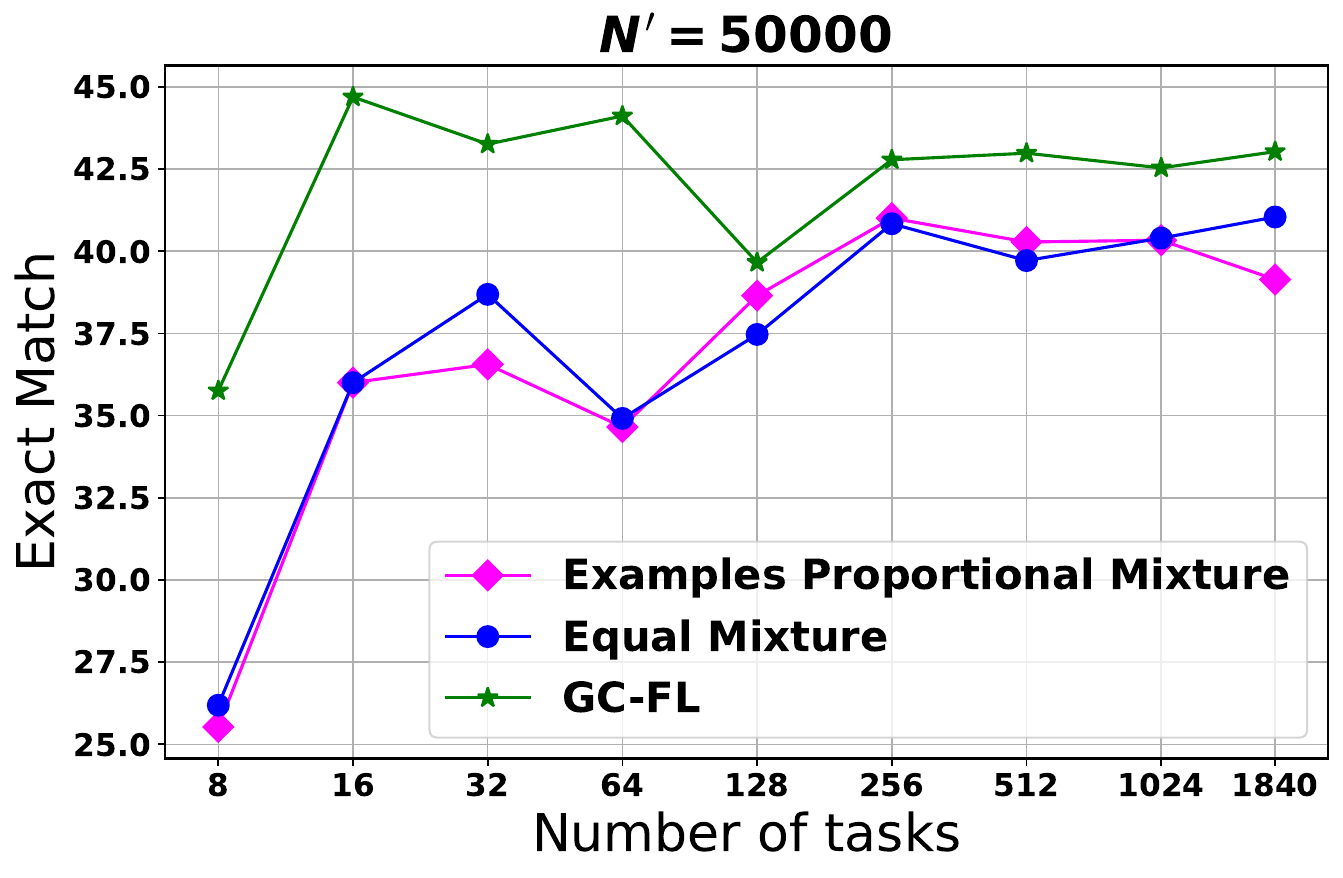}
    \end{subfigure}
    \begin{subfigure}[b]{0.32\textwidth}
        \includegraphics[width=\linewidth]{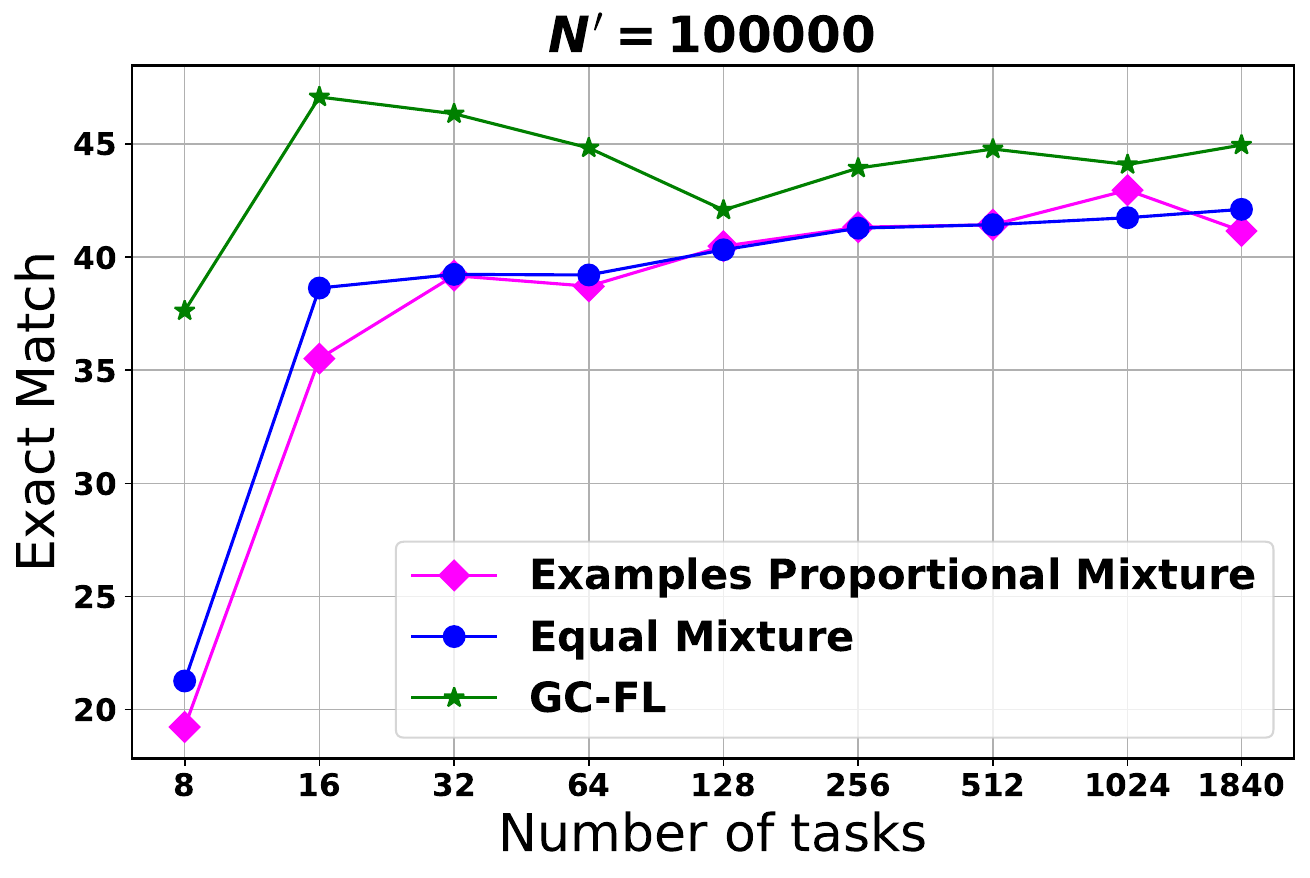}
    \end{subfigure}
    \begin{subfigure}[b]{0.32\textwidth}
        \includegraphics[width=\linewidth]{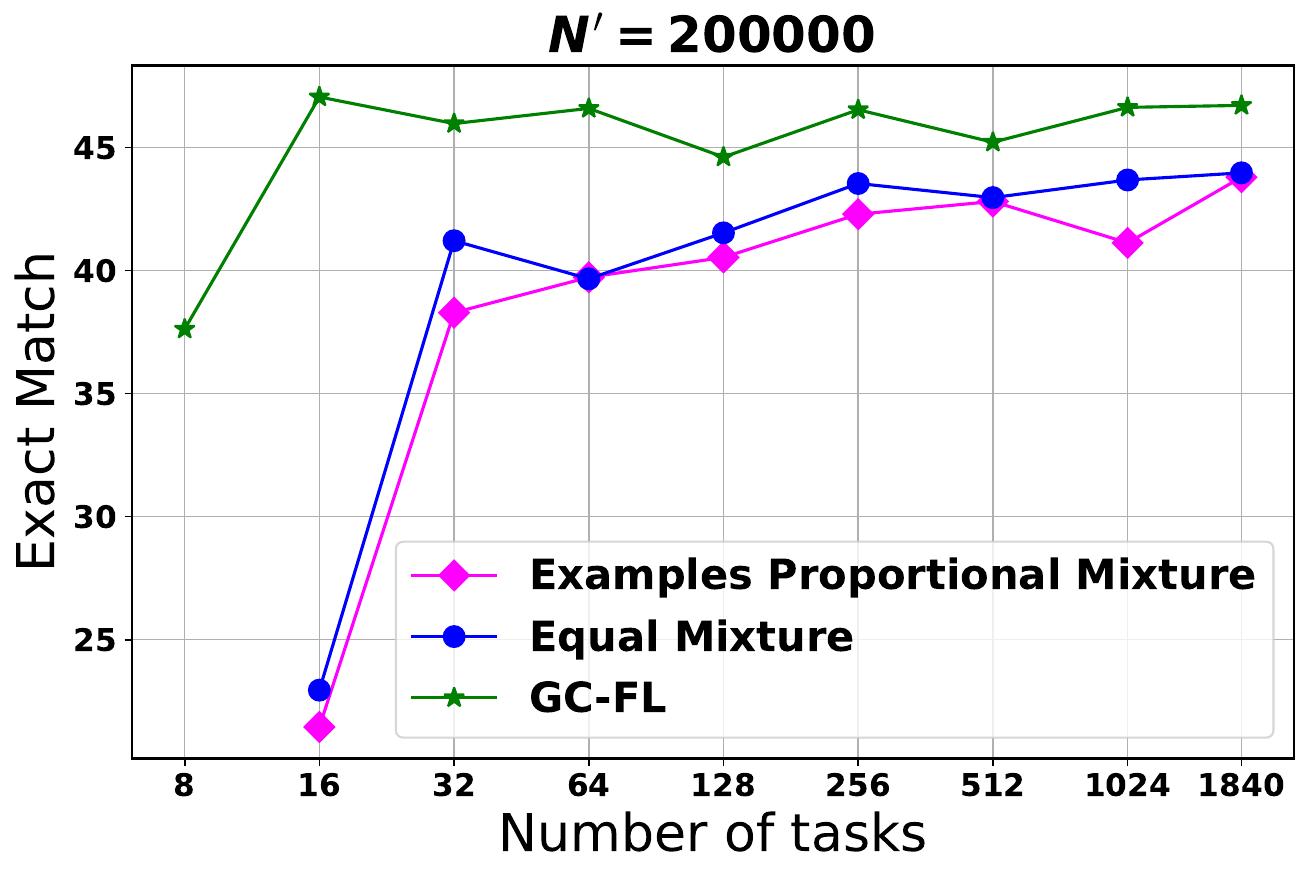}
    \end{subfigure}
    \begin{subfigure}[b]{0.32\textwidth}
        \includegraphics[width=\linewidth]{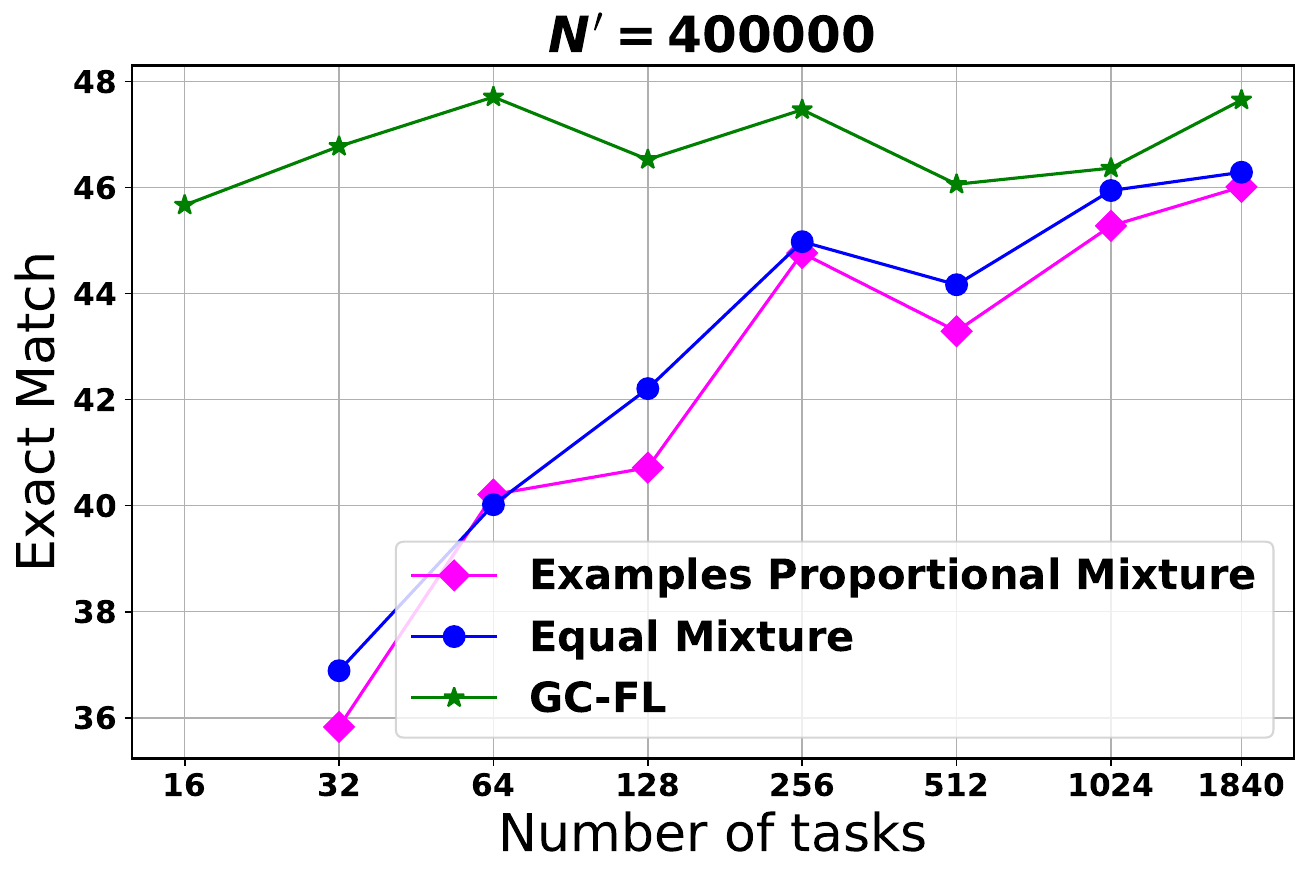}
    \end{subfigure}
    \caption{Task-Scaling Curves in the case of Llama-2-7b for different $N'$s, where X-axis is the number of tasks ($M'$) and Y-axis is the weighted average of Exact Matches on MMLU and BBH.}
    \label{fig:llama-2_task_scaling}
\end{figure*}

\subsection{Addressing \textbf{RQ2} ($M'<M$)}
\label{sec:addressing_rq2}

\textbf{RQ2} is an instance of the subset selection problem defined in Section~\ref{sec:approach}, where $M'<M$ i.e., studying the effect of pruning some tasks altogether, on both \model{} and the baseline data mixtures. We fine-tune Llama-2-7b on both the baseline and \model{} data mixtures by varying the number of tasks ($M'$) in $\{8, 16, 32, 64, 128, 256, 512, 1024, 1840\}$ and the total number of instances ($N'$) in $\{25000, 50000, 100000, 200000, 400000\}$. Figure~\ref{fig:llama-2_task_scaling} depicts the task-scaling plots for each $N'$. Baseline data mixtures' performance steadily improves upon increasing the number of tasks, even though the gains are diminishing after a point. In contrast, \model{} data mixtures yield optimal performance at an in-between point and this performance does not seem to sustain upon adding more and more tasks, suggesting that rather than increasing the tasks, focusing on representative tasks and sampling more instances from these might be more beneficial in low-budget scenarios. Even with an ample budget, scaling tasks should be done judiciously, as close to ~97\% of performance achievable by using the entire FLAN collection can be attained with just a subset of 16 representative tasks alone with $N'=200000$.

\begin{table}[b!]
    \centering
    % \resizebox{0.92\linewidth}{!}{
    \begin{tabular}{lcc}
        \toprule
        $N'$ & FLANv2 Mix & \model{} \\
        \midrule
        $25000$ & \textbf{40.05} & 39.8 \\
        $50000$ & 41.49 & \textbf{43.03} \\
        $100000$ & 43.18 & \textbf{44.96} \\
        $200000$ & 44.73 & \textbf{46.7} \\
        $400000$ & 46.26 & \textbf{47.65} \\
        \hline
    \end{tabular}
    % }
    \caption{Comparison of \model{} with mixture weights of \citet{longpre2023flan} for Llama-2-7b. The scores are weighted average of exact match on MMLU and BBH.}
    \label{tab:flan_v2_comparison}
\end{table}

\subsection{Comparison with FLANv2 Mix}
\label{sec:comparison_with_flan_v2_mix}

We now compare \model{} with FLANv2-mix i.e., by using the mixture weights suggested by \citet{longpre2023flan} where the weights 40\%, 32\%, 20\%, 5\% and 3\%  are assigned to the 5 sub mixtures of FLAN 2022 --- FLANv1, T0, NIV2, CoT and Dialog respectively and instances are  randomly sampled from each sub mixture. Since this method only prescribes weights on sub mixtures and not on individual tasks, we only consider the case where $M'=M$. Table~\ref{tab:flan_v2_comparison} contains comparison of \model{} vs FLANv2-mix for Llama-2-7b. FLANv2-mix seems to perform better than EPM (Baseline-1) and EM (Baseline-2) in some cases although \model{} almost always outperforms FLANv2-mix.

\subsection{Ablation Study: Submodular Function}
\label{sec:ablation_submodular_function}

\begin{table}[b!]
    \centering
    \resizebox{\linewidth}{!}{
        \begin{tabular}{cc}
            $N'=25000$ & $N'=50000$ \\
            \begin{tabular}{c|ccc}
                \multicolumn{1}{r|}{\diagbox{$f_1$}{$f_2$}} & FL & GC & LOGDET \\
                \hline
                FL & 28.81 & 27.02 & 28.05 \\
                GC & 39.8 & 40.84 & 40.61 \\
                LOGDET & 38.83 & 35.52 & \textbf{41.38} \\
            \end{tabular}
            & 
            \begin{tabular}{c|ccc}
                \multicolumn{1}{r|}{\diagbox{$f_1$}{$f_2$}} & FL & GC & LOGDET \\
                \hline
                FL & 25.31 & 25.95 & 28.35 \\
                GC & \textbf{43.03} & 40.98 & 42.28 \\
                LOGDET & 41.97 & 40.67 & 41.84 \\
            \end{tabular} \\
            & \\
            $N'=10000$ & $N'=200000$ \\
            \begin{tabular}{c|ccc}
                \multicolumn{1}{r|}{\diagbox{$f_1$}{$f_2$}} & FL & GC & LOGDET \\
                \hline
                FL & 25.42 & 25.2 & 26.89 \\
                GC & \textbf{44.96} & 43.92 & 43.61 \\
                LOGDET & 43.63 & 42.25 & 42.81 \\
            \end{tabular}
            & 
            \begin{tabular}{c|ccc}
                \multicolumn{1}{r|}{\diagbox{$f_1$}{$f_2$}} & FL & GC & LOGDET \\
                \hline
                FL & 43.58 & 41.67 & 42.72 \\
                GC & \textbf{46.7} & 45.91 & 46.21 \\
                LOGDET & 43.75 & 43.86 & 43.82 \\
            \end{tabular} \\
        \end{tabular}
    }
    \caption{Grid Search on submodular functions $f_1$ and $f_2$ for Llama-2-7b where weighted average of exact match on MMLU and BBH are compared for different choices of $f_1$ and $f_2$. FL denotes Facility Location, GC denotes Graph Cut and LOGDET denotes Log-Determinant.}
    \label{tab:grid_search_submodular_functions}
\end{table}

\begin{figure}[h!]
    \centering
    \includegraphics[width=0.75\linewidth]{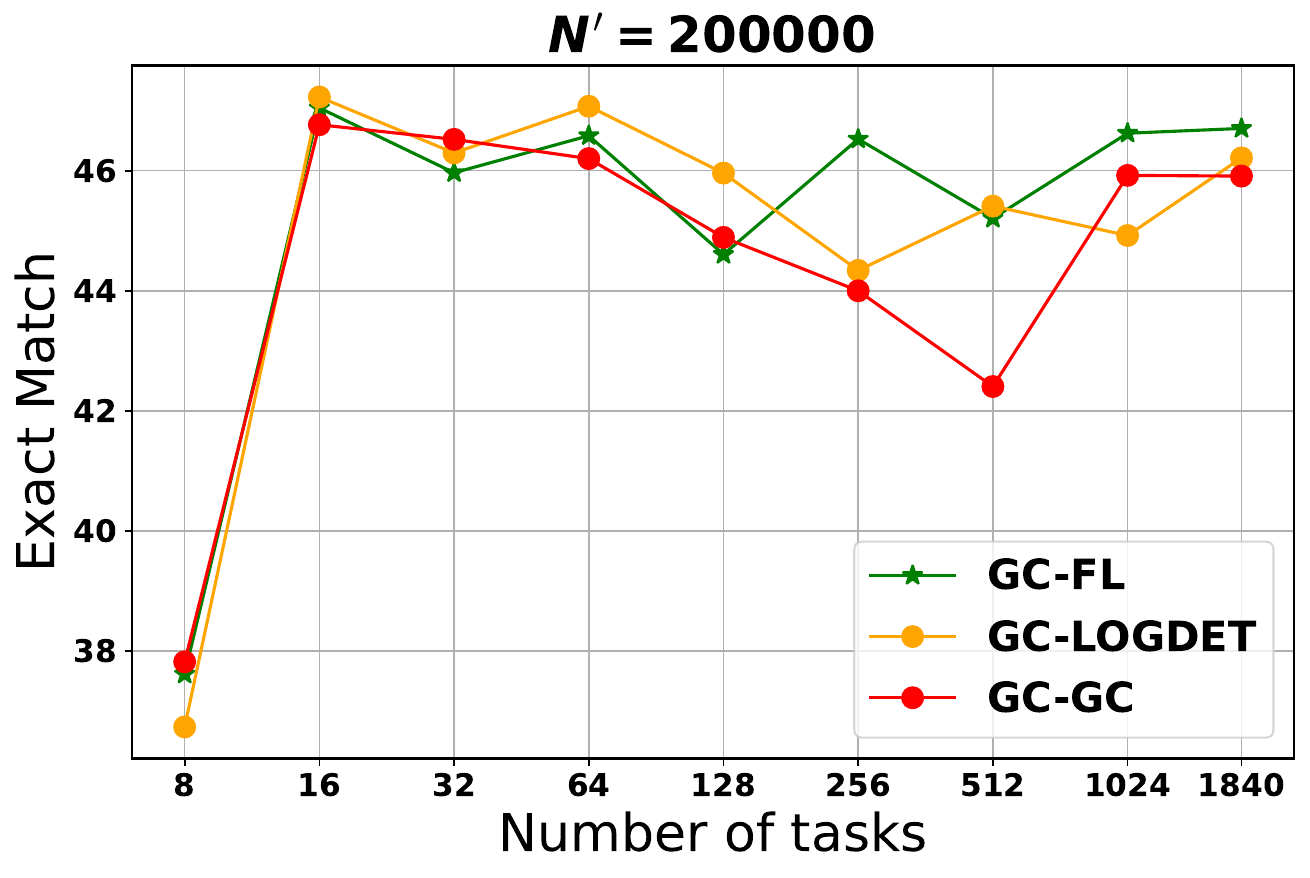}
    \caption{Varying $f_2$ when $f_1$ is Graph Cut}
    \label{fig:ablation_gc_as_f1}
\end{figure}

Both Stage-1 and Stage-2 of \model{} require us to choose submodular functions $f_1$ and $f_2$ - which are best treated as hyperparameters because of uncertainty with respect to which functions are best suited for instruction tuning (Section~\ref{sec:choosing_f1_f2}). We conduct a grid search on $f_1$ and $f_2$ using three functions from Table~\ref{tab:submodular_functions_examples}, by setting $M'=M$ and varying $N'$ in $\{25000, 50000, 100000, 200000\}$. In this work, we use $\lambda=0.4$ for Graph-Cut. The grid search (summarized in Table~\ref{tab:grid_search_submodular_functions}) suggests that Graph Cut is optimal for task subset selection, while Facility Location is best for instance selection when $M'=M$. The task scaling curves for each combination of $f_1$ and $f_2$ (for different $N'$s) are present in Appendix~\ref{app:task_scaling_curves_f1_f2}. However, in this section, we highlight the case where $f_1$ is Graph Cut and $f_2$ is varied. We find that optimal $f_2$ in this case also depends on $M'$. For instance, in Figure~\ref{fig:ablation_gc_as_f1} where $N'=200000$, Facility Location performs the best at higher $M'$, highlighting the importance of representation, while Graph Cut and Log Determinant show better performance at lower $M'$, highlighting the importance of diversity. We hypothesize this is because --- for higher $M'$s, each task on average gets a relatively low budget and the need for representation dominates the need for diversity; however when there is sufficient enough budget for each task, i.e., for lower $M'$s, the need for diversity takes over.

\subsection{Ablation Study: LLM}
\label{sec:ablation_llm}

We also test the effectiveness of \model{} strategy on two other LLMs - Falcon-7B and Mistral-7B. Table~\ref{tab:ablation_llm} presents a comparison of \model{} mixture with the baseline mixtures for Falcon and Mistral when $M'=M$. For Falcon, we only report MMLU-zero-shot since Falcon gets an exact match of 0.0 for BBH-zero-shot even after fine-tuning. The task-scaling curves for these models are present in Appendix~\ref{app:task_scaling_curves_mistral_falcon}.

\begin{table}[h!]
    \centering
    \resizebox{0.92\linewidth}{!}{
    \begin{tabular}{lccc}
        \toprule
        $N'$ & Data Mix. & Mistral & Falcon$^*$ \\
        \midrule
        \multirow{4}{*}{$25000$} & EPM(Baseline-1) & 52.69 & 23.09 \\
         & EM(Baseline-2) & 51.65 & 21.1 \\
         & FLANv2-mix & 53.24 &  \textbf{26.16} \\
         & \model{} & \textbf{53.33} & 22.73 \\
        \hline
        \multirow{4}{*}{$50000$} & EPM(Baseline-1) & 53.75 & 19.86 \\
         & EM(Baseline-2) & 53.16 & 22.96 \\
         & FLANv2-mix & \textbf{54.15} & 25.66  \\
         & \model{} & 53.8 & \textbf{26.31} \\
        \hline
        \multirow{4}{*}{$100000$} & EPM(Baseline-1) & 54.24 & 20.88 \\
         & EM(Baseline-2) & \textbf{54.57} & 24.85 \\
         & FLANv2-mix & 54.14 & \textbf{27.47} \\
         & \model{} & 53.75 & 27.22 \\
        \hline
        \multirow{4}{*}{$200000$} & EPM(Baseline-1) & 55.25 & 24.7 \\
         & EM(Baseline-2) & 54.71 & 24.39 \\
         & FLANv2-mix & 55.27 & 27.63 \\
         & \model{} & \textbf{55.4} & \textbf{30.96} \\
        \hline
    \end{tabular}
    }
    \caption{Comparison of \model{} with baselines for Mistral and Falcon. The scores correspond to weighted averages of exact match on MMLU and BBH.}
    \label{tab:ablation_llm}
\end{table}

\subsection{Discussion}
\label{sec:insights_and_discussions}

In Section~\ref{sec:ablation_submodular_function}, we saw that Graph Cut proves to be the most effective for selecting the weighted task subsets. Figures~\ref{fig:fl_visualization},~\ref{fig:gc_visualization},~\ref{fig:logdet_visualization} of Appendix~\ref{app:task_subsets_visualization} contain t-SNE visualizations for task subsets selected by the three submodular functions listed in Table~\ref{tab:submodular_functions_examples}. Facility Location being a sum-max formulation, a single point is sufficient to represent a cluster and hence it gives more weightage to cluster centers and very less weightage to others. As a result, the submodular gains for first few selected points are very high and then the gains quickly become very small for other points in case of facility location. Log Determinant on the other hand predominantly selects diverse points not taking representation into account. Graph Cut, which models both, hence performs better than both Facility Location and Log Determinant for selecting tasks. 

Further, in Figure~\ref{fig:llama-2_task_scaling} of Section~\ref{sec:addressing_rq2} we saw that, at around 16 tasks, there is a peak in performance, after which there is a slight decline. To facilitate more investigation into what tasks are assigned more weightage, Table~\ref{tab:gains_table} and Table~\ref{tab:gains_table_last_128} list down the top-128 and last-128 tasks in the submodular ordering obtained using graph cut. While the former set contains more traditional NLP tasks like Natural Language Inference, Next Sentence Prediction, Question Answering, Summarization, {\em etc .}, the latter set mostly consists of tasks like Program Execution.

\section{Related Work}
\label{sec:related_work}

\subsection{Data for Instruction Tuning}
Following \citet{wang2023data}, we summarize the related work in three categories --- data quality, data quantity, and task composition.

\paragraph{Data Quantity}
Research diverges on scaling instruction data quantity, with some advocating for limited data~\citep{zhou2023lima,chen2023maybe} to expose pretraining knowledge, while others argue for scaling up~\citep{wei2021finetuned,sanh2021multitask}. According to \citet{ji2023exploring,dong2023abilities,yuan2023scaling,song2023dynamics}, the impact of scaling varies across tasks and model abilities.~\citet{alshikh2023becoming} also introduce a metric for instruction following ability, and suggest an early stopping criterion for instruction-tuning.

\paragraph{Data Quality} High-quality data is crucial in instruction tuning~\citep{chia2023instructeval,ding2023enhancing,zhou2023lima}. \citet{wang2023harnessing} use perplexity to select suitable instructions generated by models, while \citet{li2023self} employ the language model itself to augment and curate high-quality training examples to improve its own performance. \citet{cao2023instruction} propose InstructionMining, utilizing natural language indicators to predict inference loss as a proxy for data quality without human intervention and select the best subset based on this. \citet{chen2023alpagasus} introduce AlpaGasus, which uses a strong LLM to select high-quality subsets.
\citet{lu2023self} and \citet{madaan2024self} also leverage the power of fine-tuned LLM itself to evaluate the quality of instructions.~\citet{attendu2023nlu} employs dynamic data subset selection by filtering out unimportant samples \textit{during} finetuning, based on an extended EL2N metric~\citep{paul2021deep,fayyaz2022bert}. \citet{taori2023stanford} propose \#InsTag to assess instruction diversity in SFT datasets using ChatGPT. Additionally, \citet{wan2023explore} propose Explore-Instruct, utilizing LLMs to actively explore domain-specific spaces and gather diverse instruction-tuning data. \citet{wu2023self} select new data points that are distinct from existing ones in the model embedding space, augmenting the training dataset iteratively to enhance diversity within subsets.

\paragraph{Task Composition}
Many previous works show the benefit of scaling the number of tasks~\citep{wei2021finetuned,chung2022scaling,wang2022super,sanh2021multitask}. However, works like \citet{iyer2022opt,longpre2023flan} have also acknowledged that task balancing is also very important for effective instruction tuning. \citet{dong2023abilities} explore data composition across GSM8k, Code Alpaca, and ShareGPT datasets, finding differential impacts of data scaling on performance across abilities and propose a Dual-stage mixed fine-tuning strategy as a promising solution to activate multiple abilities efficiently. \citet{ivison2022data} identifies relevant multitask subsets based on the similarity between the pre-trained model's representations, using a small amount of target task data. \citet{yin2023dynosaur} uses instruction representations for task selection, acting as a replay strategy to mitigate catastrophic forgetting and improve generalization in continual learning. \citet{yue2023mammoth} construct math generalist models via instruction tuning on hybrid of chain-of-thought and program-of-thought rationales in math.~\citet{lou2023prompt,zhang2023instruction} provide a survey of instruction tuning in general and \citet{wang2023data} provide a detailed survey of data management for instruction tuning. 

\subsection{Submodularity for Subset Selection}
Submodularity~\citep{fujishige2005submodular} has a long history in combinatorial optimization, game theory, economics etc~\citep{edmonds1970matroids,lovasz1983submodular,carter2001foundations,topkis1998supermodularity}. It has recently gained traction in machine learning where it has been used for data subset selection for machine translation~\citep{kirchhoff2014submodularity}, speech recognition~\citep{wei2014unsupervised,mittal2022partitioned,kothawade2023ditto}, efficient pre-training of language models~\citep{ingenious}, active learning~\citep{wei2015submodularity,kothawade2021similar}, hyperparameter tuning~\citep{killamsetty2022automata}, domain adaptation~\citep{karanam2022orient}, computer vision~\citep{kaushal2019learning}, continual learning~\citep{tiwari2022gcr} etc. For a more detailed review of submodularity and its applications, please refer to the survey by \citet{bilmes2022submodularity}.

\newpage
\section{Conclusion \& Future Work}
In this paper, we introduced \model{} --- a novel data mixture strategy for instruction tuning that utilizes a submodular function to assign importance scores to tasks, determine the mixture weights, and also select non-redundant samples from each task. Further, we also reveal that in a low-budget setting, splitting the budget among a small subset of representative tasks yields superior performance when compared to dividing it among all tasks, which suggests that task scaling should be done more judiciously. Future work could explore making this method more model-specific ({\em e.g.}, modify task embedding computation) and also possibly modify the approach for targeted instruction-data selection for creating expert LLMs that specialize in specific skills such as math, code etc.

\section{Limitations}
While the approach has its own advantages of computing the data mixture only once for a given dataset and using it for as many LLMs as one may wish, the approach might benefit from taking inputs from the language model as well and perform model specific instruction tuning. Secondly, \model{} shows that there is an optimal number of tasks at which there is peak in performance observed. However, it doesn't say anything on how to find the optimal point as it may depend on language model, the total budget and most importantly the underlying dataset. 

\section{Ethical Considerations}
While instruction tuning leverages pretrained language models and encompasses similar considerations, we anticipate that this approach will predominantly yield positive outcomes. It offers an enhanced methodology for task balancing, potentially allowing for more cost-effective fine-tuning of large language models compared to conventional data mixture strategies.

\section*{Acknowledgements}
The authors acknowledge the use of ChatGPT\footnote{https://chat.openai.com/} for solely paraphrasing and summarizing some parts of the paper and declare that \textit{none} of the generated content is presented in the paper without rigorous manual checking.

\newpage
\bibliography{anthology,custom}

\appendix
\appendix

\section*{APPENDIX}
\section{Code and Data}
\label{app:code_and_data}
Our code we used for instruction tuning and for creating data mixtures is open-sourced at \href{https://github.com/kowndinya-renduchintala/SMART}{https://github.com/kowndinya-renduchintala/SMART}. The code development utilized open-source tools, primarily relying on the HuggingFace library for model training with PyTorch as the underlying framework. Both PyTorch and HuggingFace are licensed under permissive licenses, with PyTorch under the BSD license and HuggingFace under the Apache 2.0 license. Additionally, submodular optimization was performed using SUBMODLIB, which is an openly accessible library on GitHub at \url{https://github.com/decile-team/submodlib} under the MIT license.

\section{Task Scaling Curves: Varying $f_1$ and $f_2$}
\label{app:task_scaling_curves_f1_f2}
\begin{figure}[b!]
    \centering
    \includegraphics[width=0.9\linewidth]{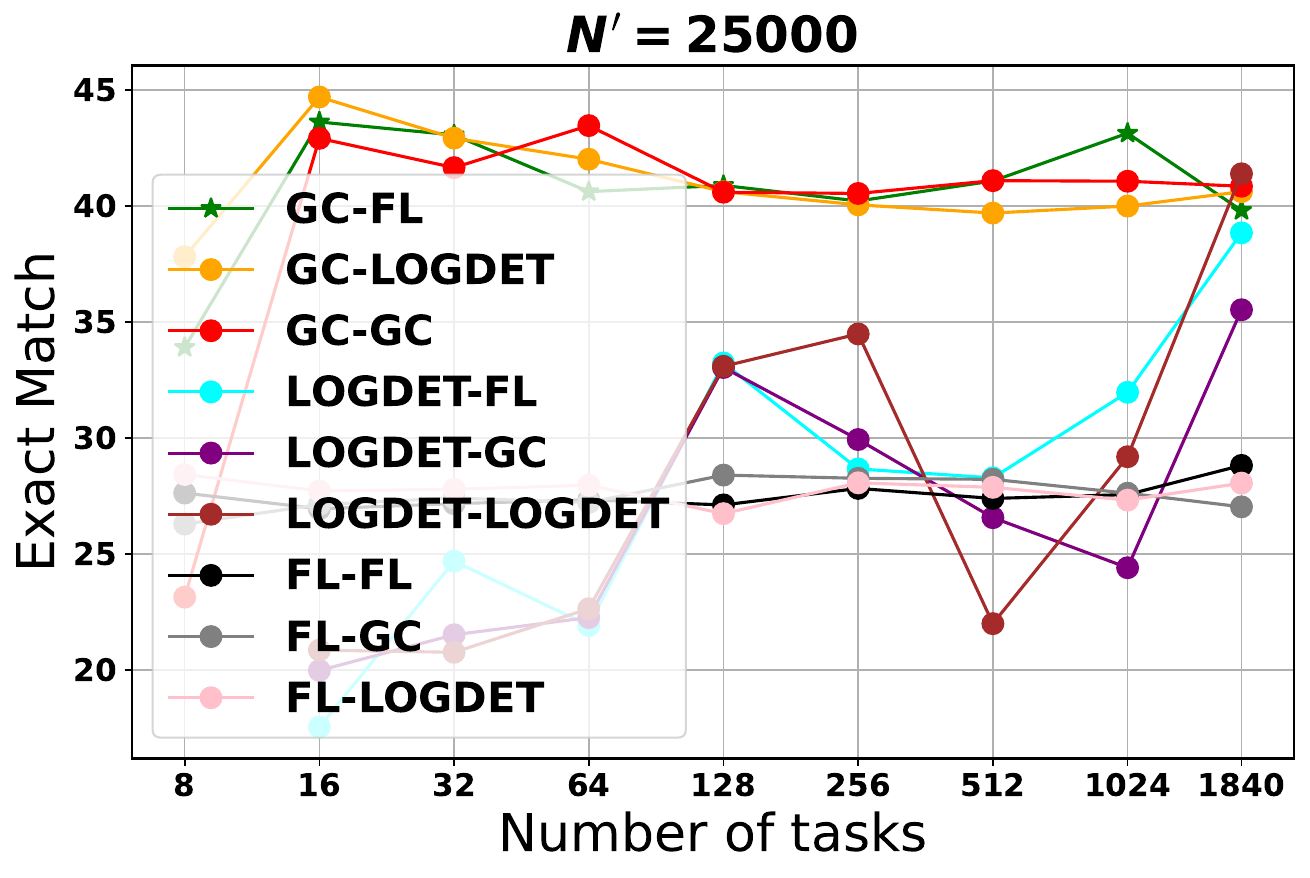}
    \includegraphics[width=0.9\linewidth]{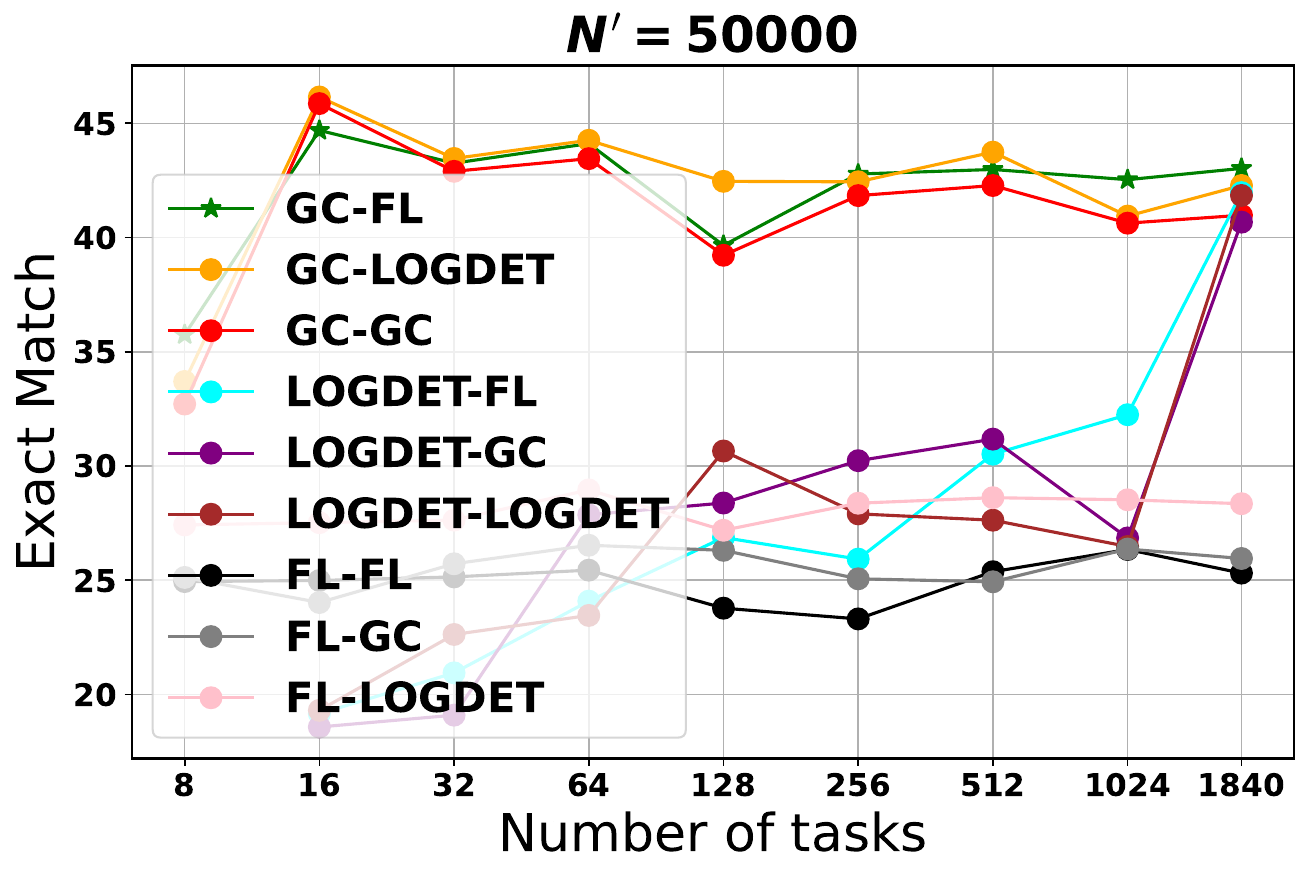}
    \includegraphics[width=0.9\linewidth]{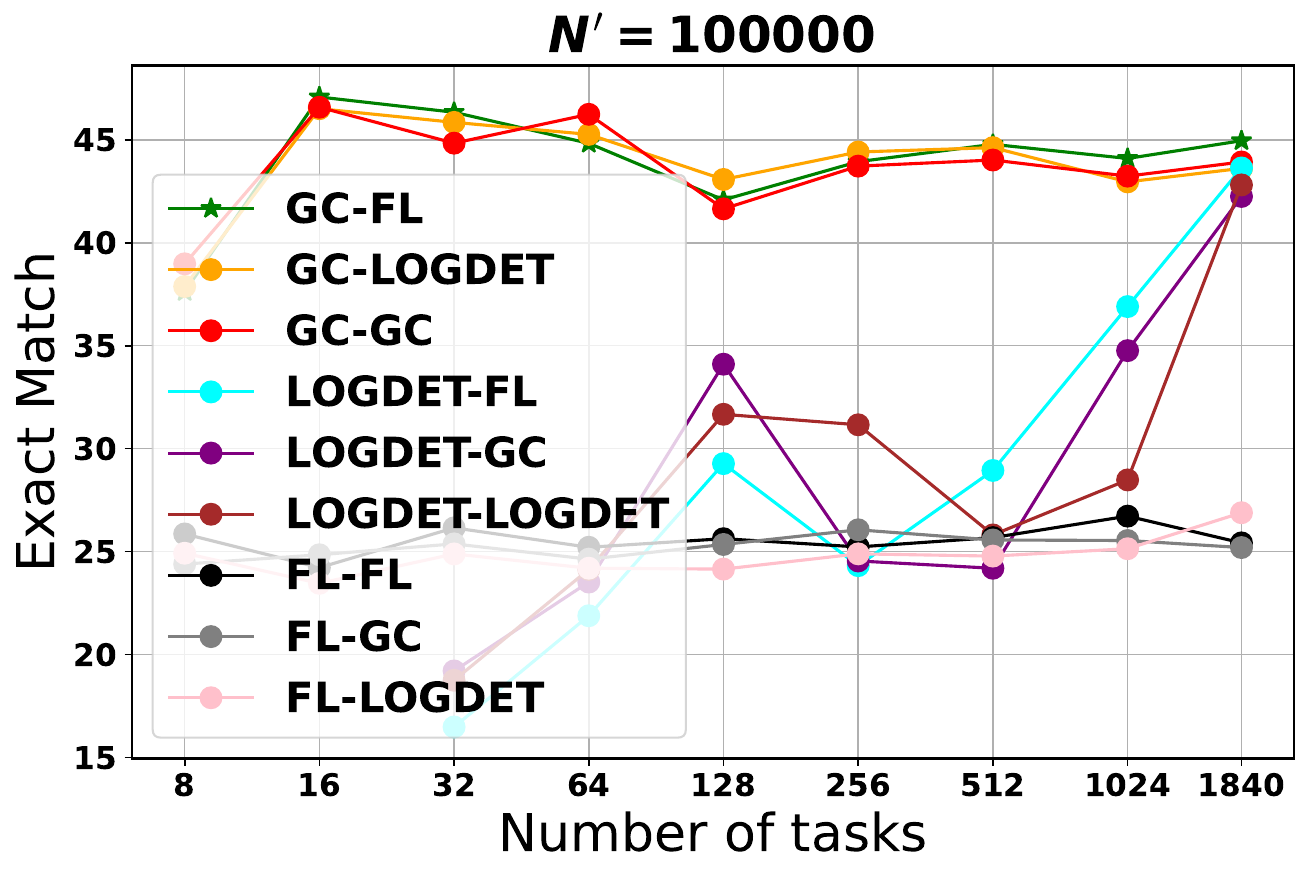}
    \includegraphics[width=0.9\linewidth]{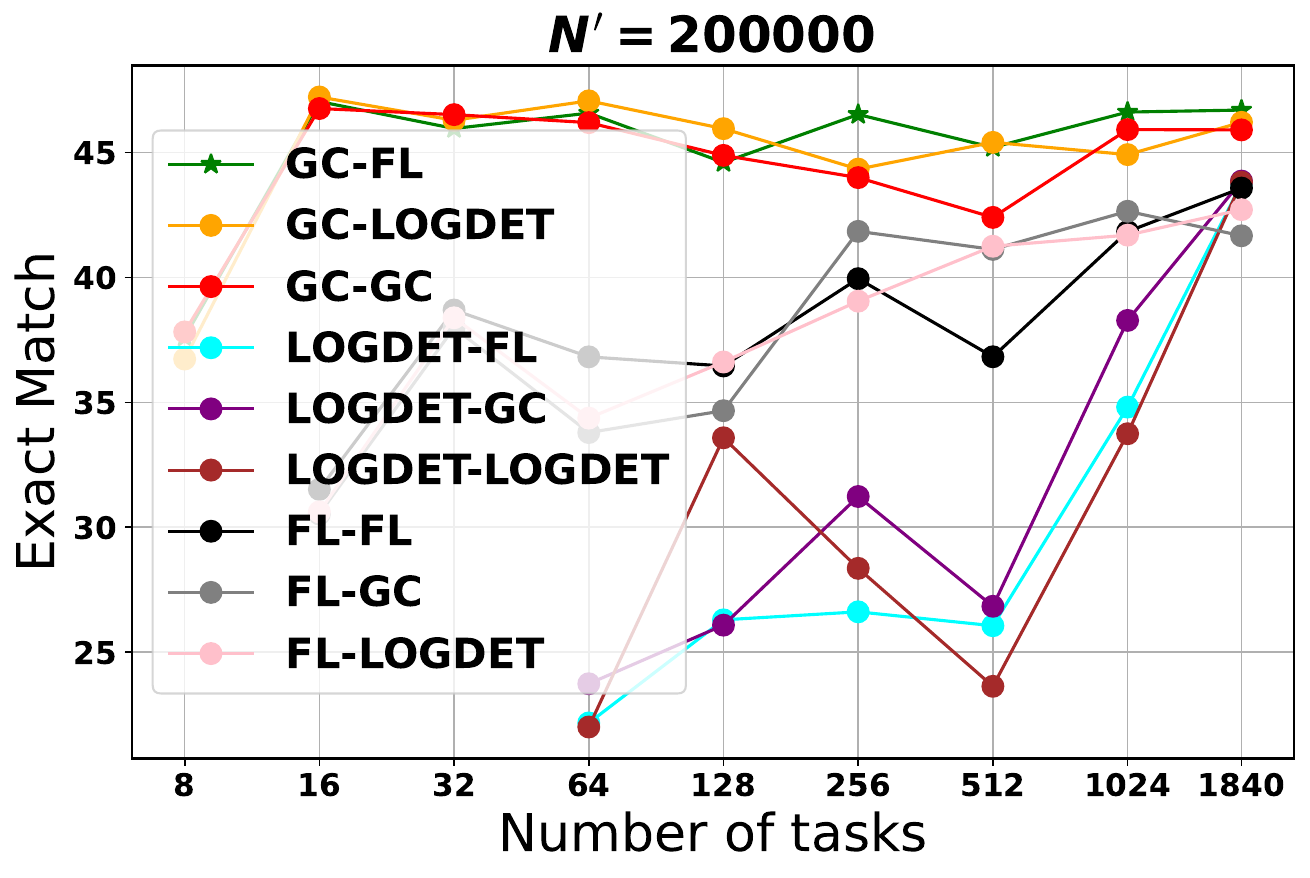}
    \caption{Task Scaling Curves for various $f_1$ and $f_2$ combinations for Llama-2-7b}
    \label{fig:ablation_f1_f2}
\end{figure}
Figure~\ref{fig:ablation_f1_f2} consists of task scaling curves for the 9 possible combinations (Section~\ref{sec:ablation_submodular_function}) of $f_1$ and $f_2$.

\section{Task Scaling Curves: Mistral, Falcon}
\label{app:task_scaling_curves_mistral_falcon}
\begin{figure}[h!]
    \centering
    \includegraphics[width=0.95\linewidth]{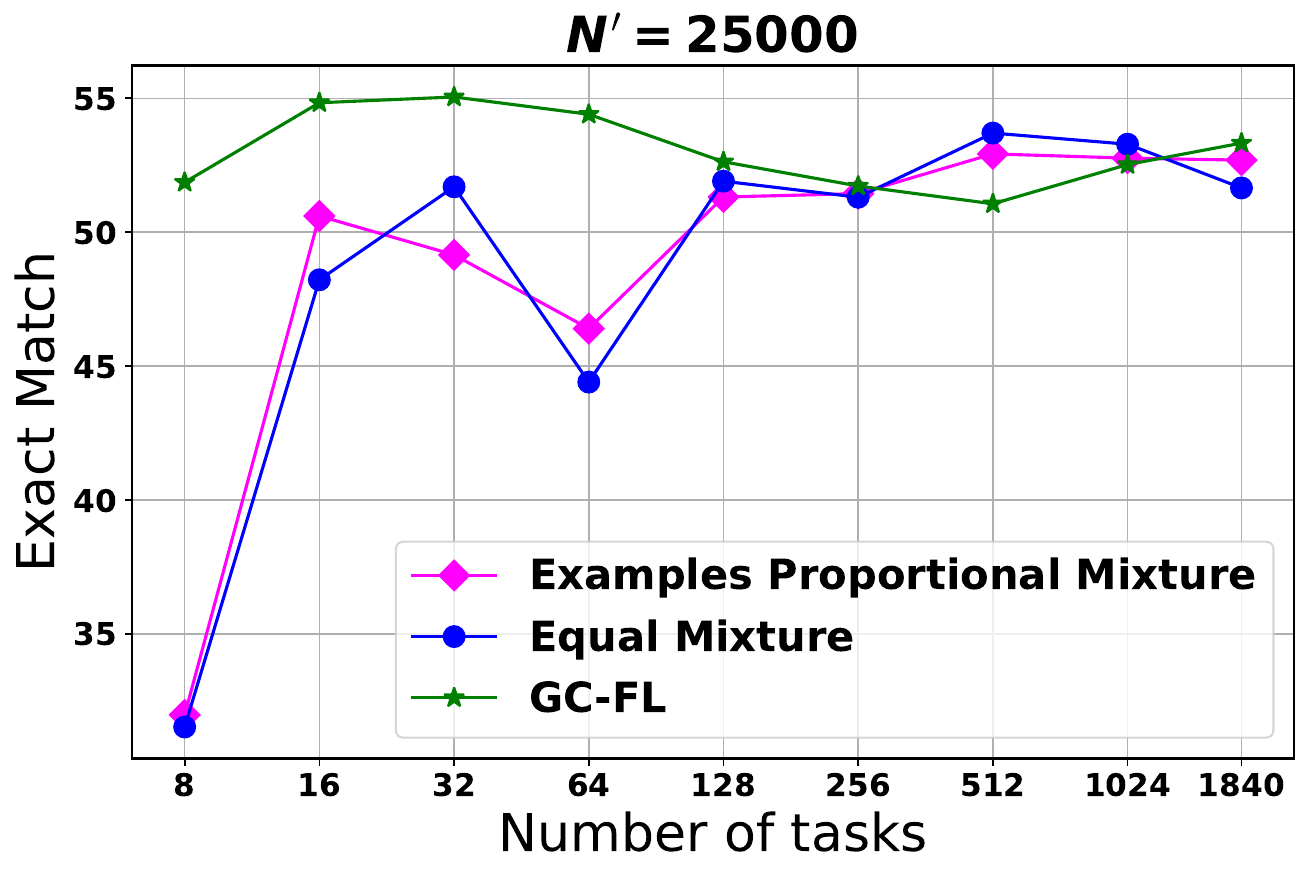}
    \includegraphics[width=0.95\linewidth]{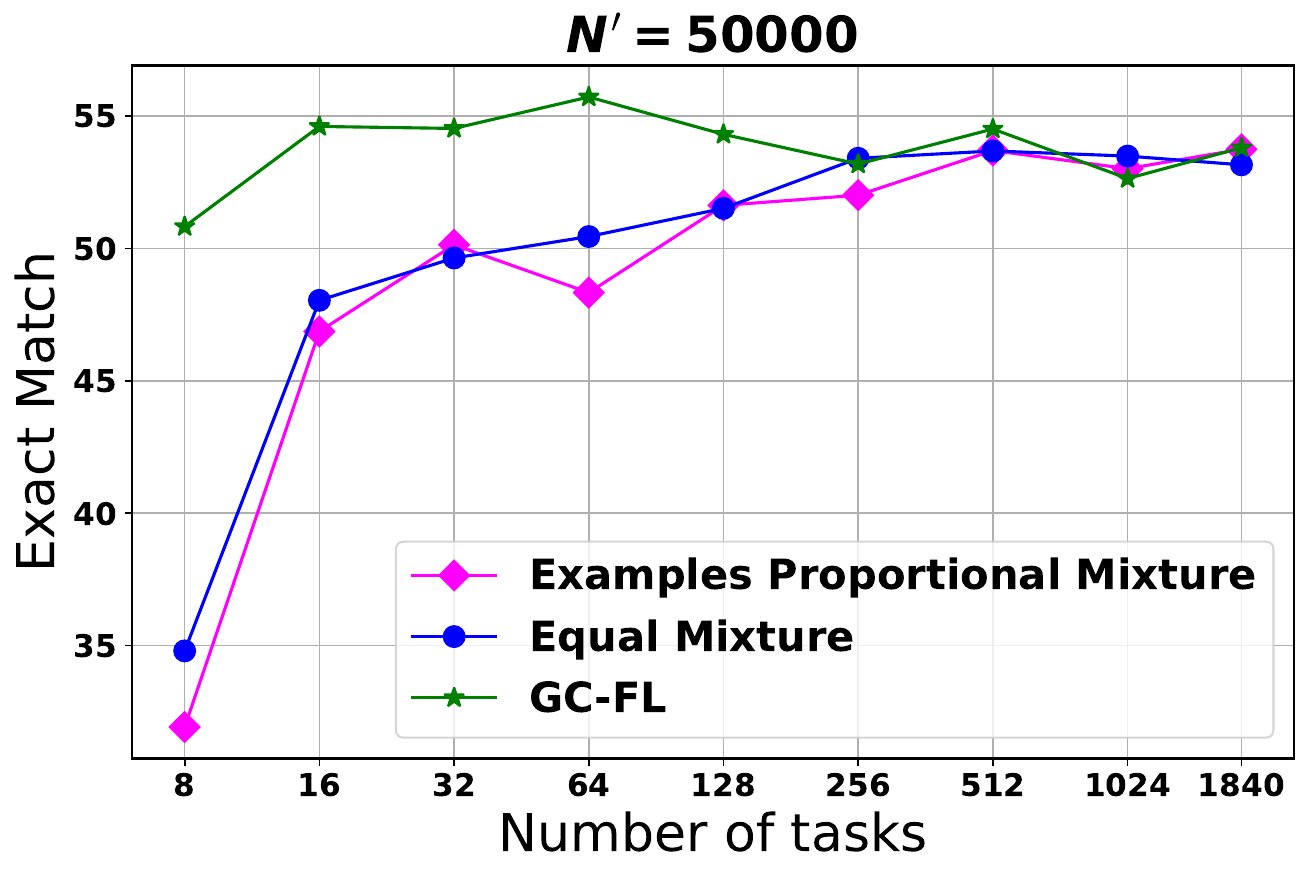}
    \includegraphics[width=0.95\linewidth]{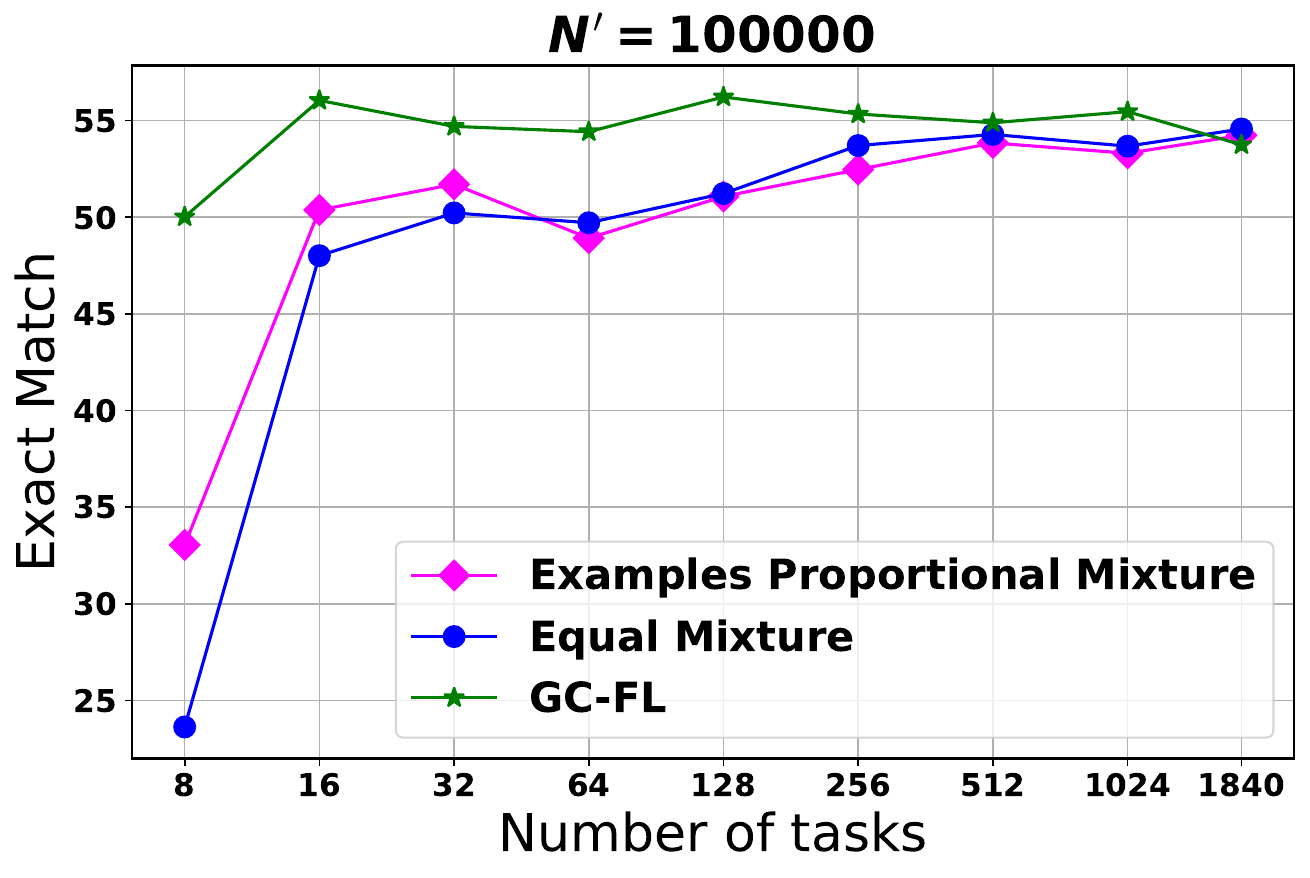}
    \includegraphics[width=0.95\linewidth]{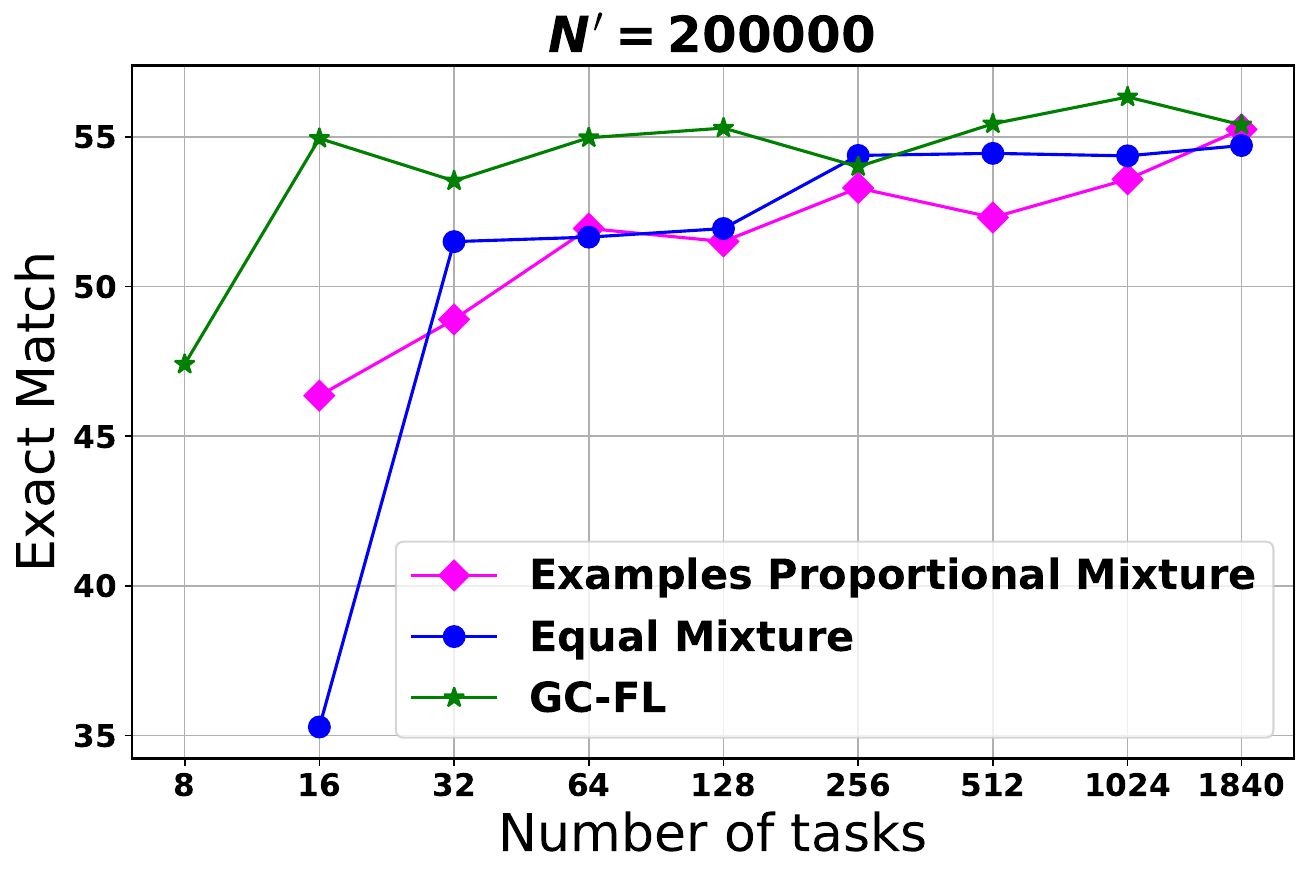}
    \caption{Task Scaling Curves for Mistral-7B}
    \label{fig:mistral_task_scaling}
\end{figure}
\begin{figure}[h!]
    \centering
    \includegraphics[width=0.95\linewidth]{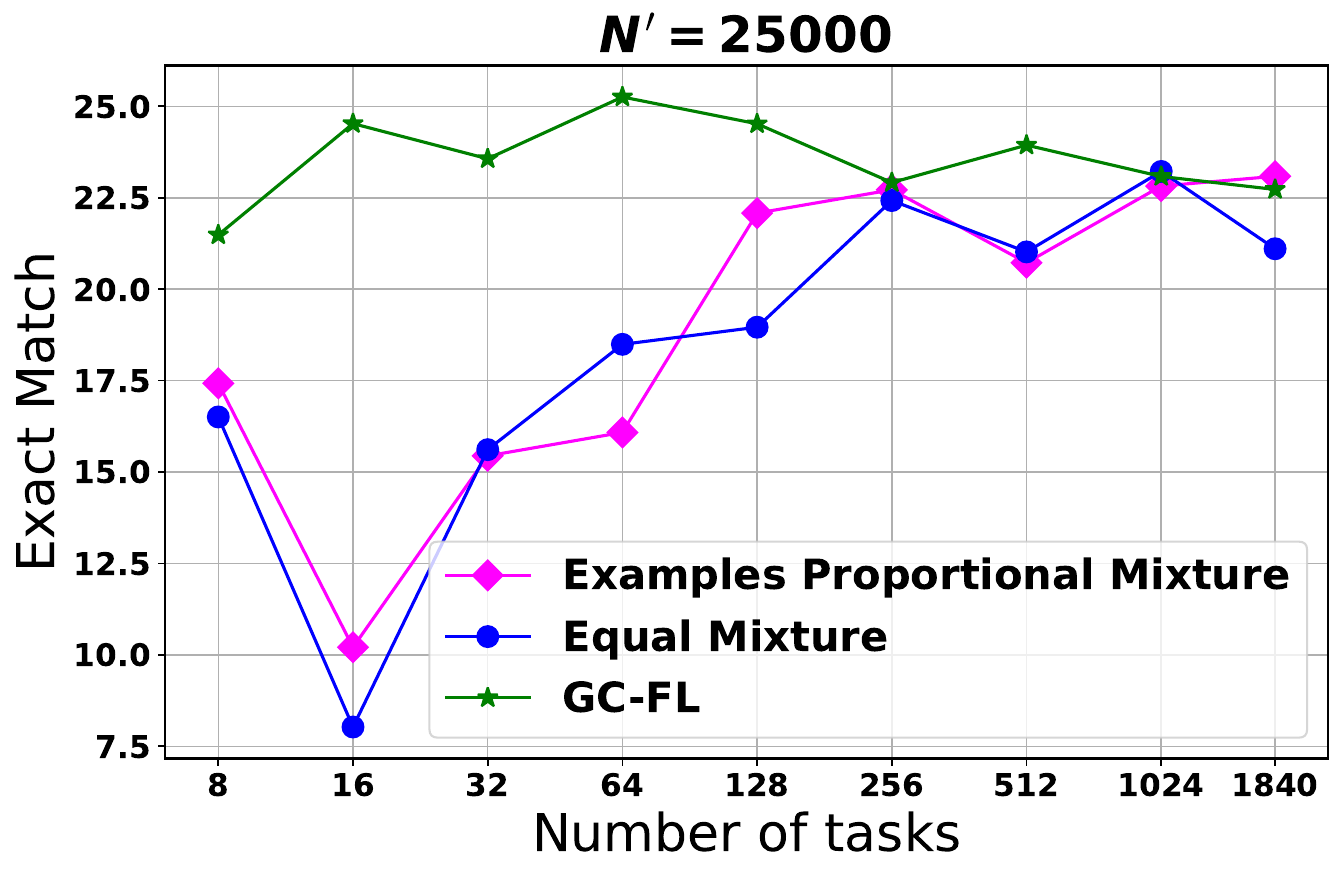}
    \includegraphics[width=0.95\linewidth]{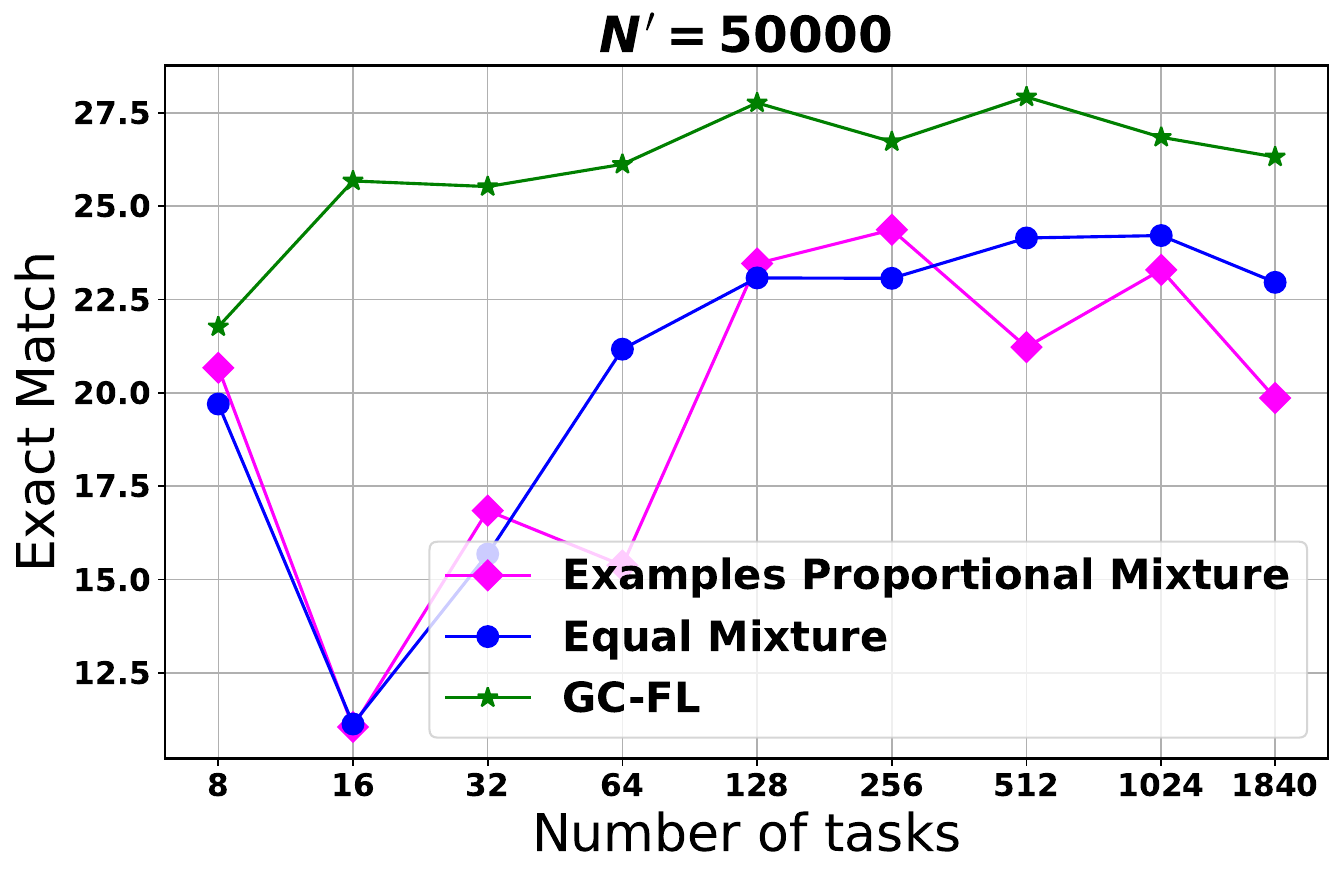}
    \includegraphics[width=0.95\linewidth]{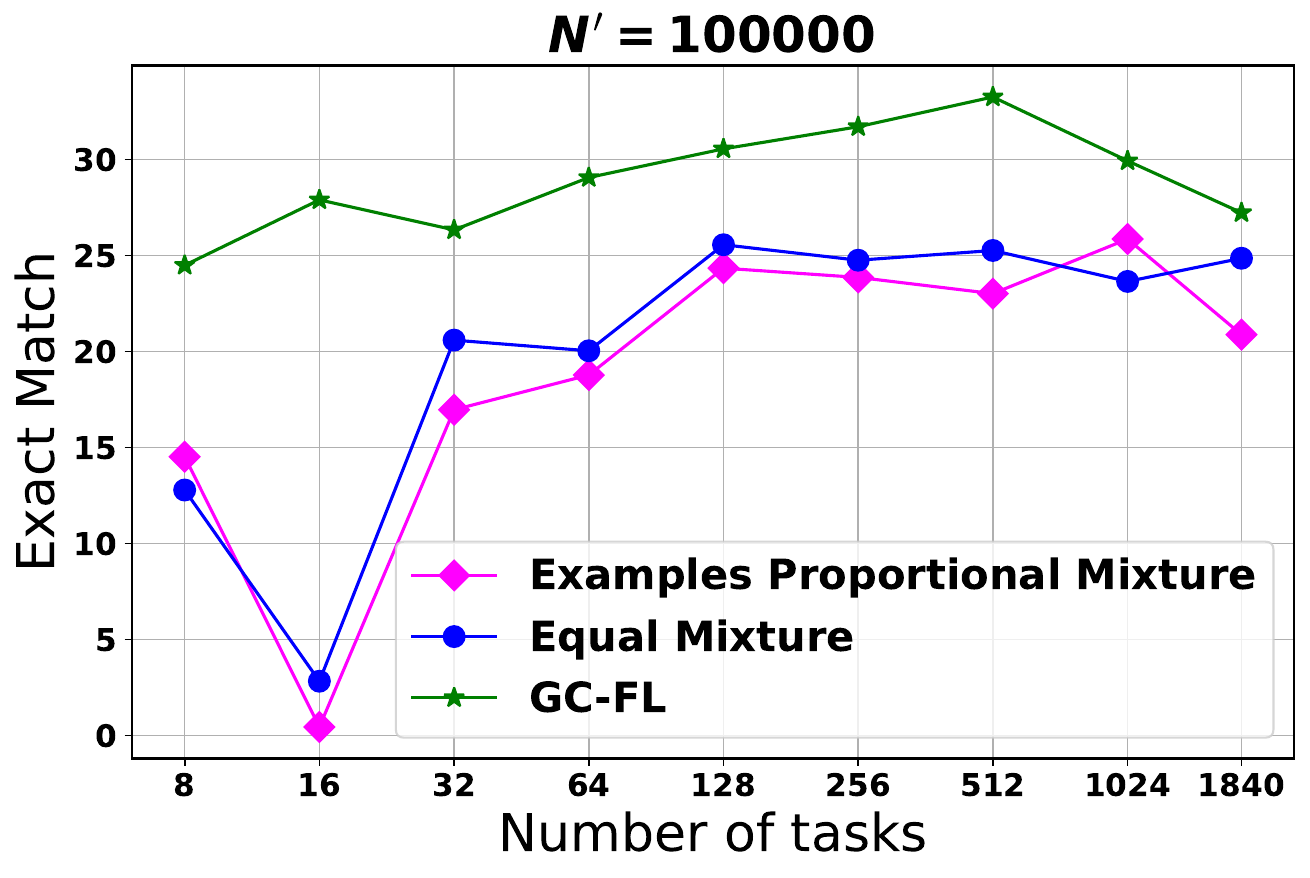}
    \includegraphics[width=0.95\linewidth]{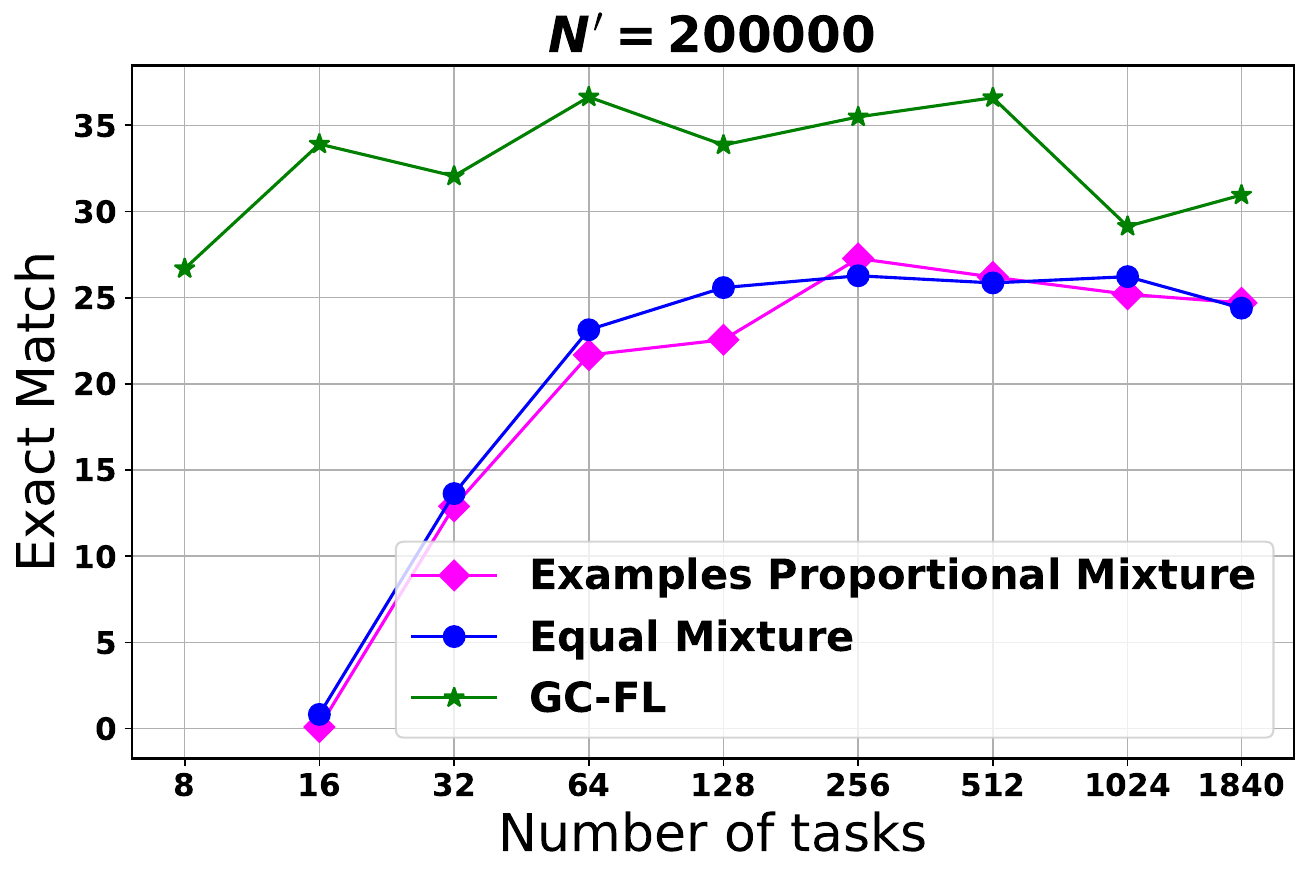}
    \caption{Task Scaling Curves for falcon-7B}
    \label{fig:falcon_task_scaling}
\end{figure}
Figure~\ref{fig:mistral_task_scaling} and Figure~\ref{fig:falcon_task_scaling} consist of task scaling curves for Mistral-7B and Falcon-7B respectively (Section~\ref{sec:ablation_llm}).

\section{Visualization of Task Subsets}
\label{app:task_subsets_visualization}
\begin{figure*}[h]
    \centering
    \begin{subfigure}[b]{0.46\linewidth}
        \centering
        \includegraphics[width=\linewidth]{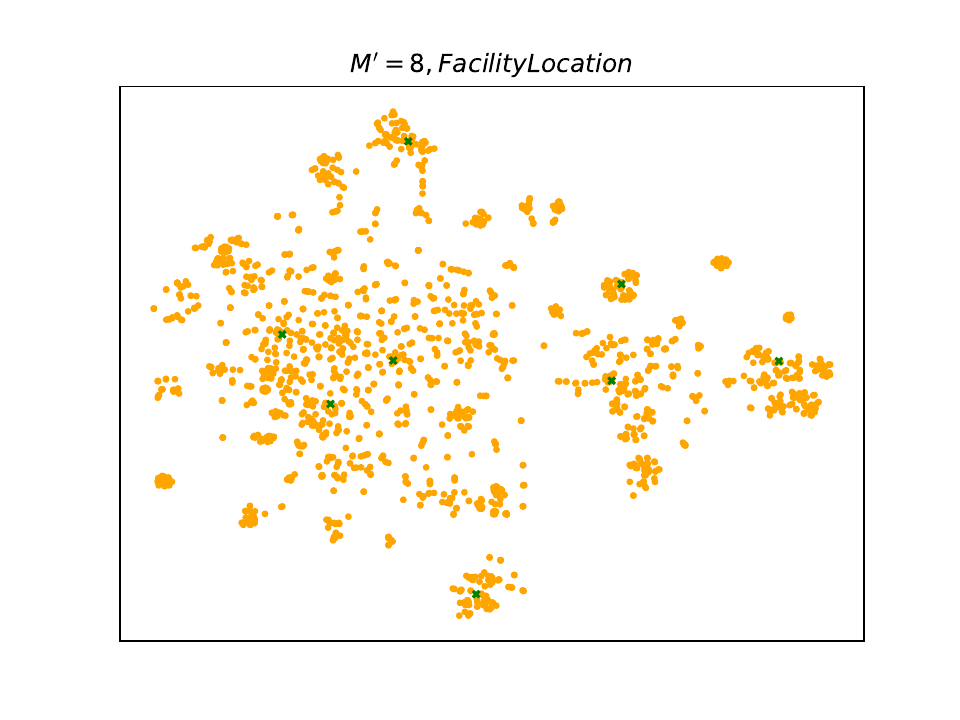}
    \end{subfigure}
    \begin{subfigure}[b]{0.46\linewidth}
        \centering
        \includegraphics[width=\linewidth]{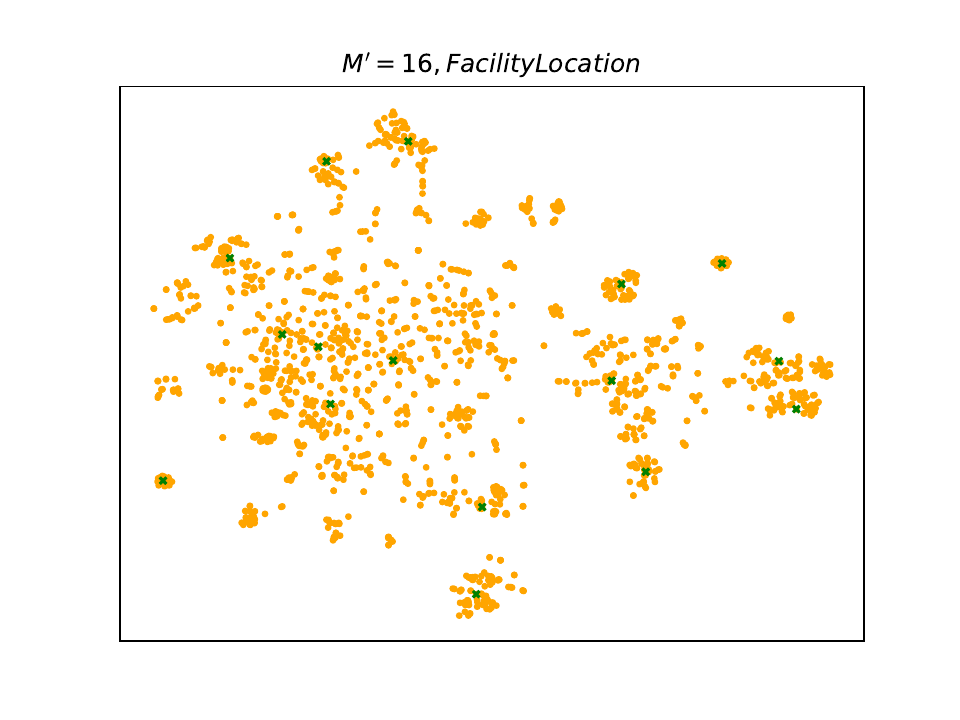}
    \end{subfigure}
    \begin{subfigure}[b]{0.46\linewidth}
        \centering
        \includegraphics[width=\linewidth]{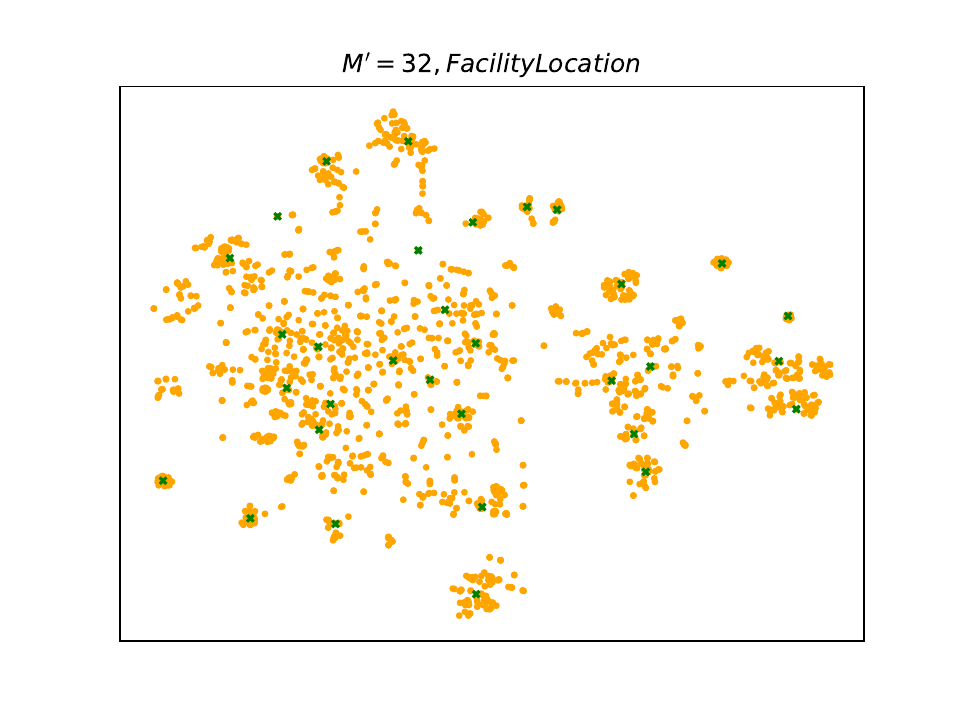}
    \end{subfigure}
    \begin{subfigure}[b]{0.46\linewidth}
        \centering
        \includegraphics[width=\linewidth]{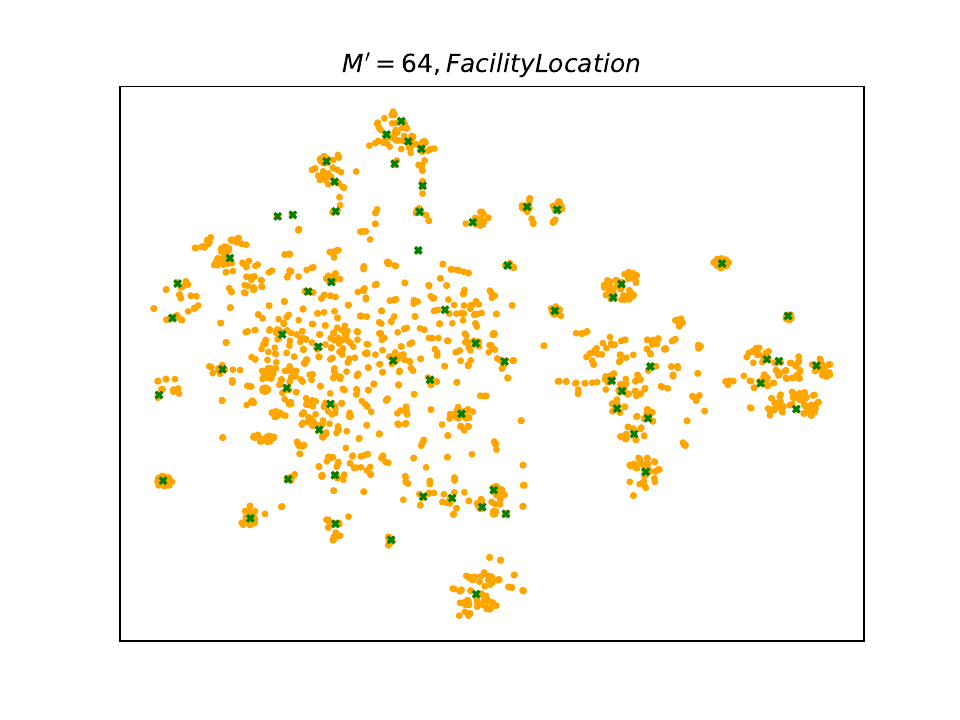}
    \end{subfigure}
    \begin{subfigure}[b]{0.46\linewidth}
        \centering
        \includegraphics[width=\linewidth]{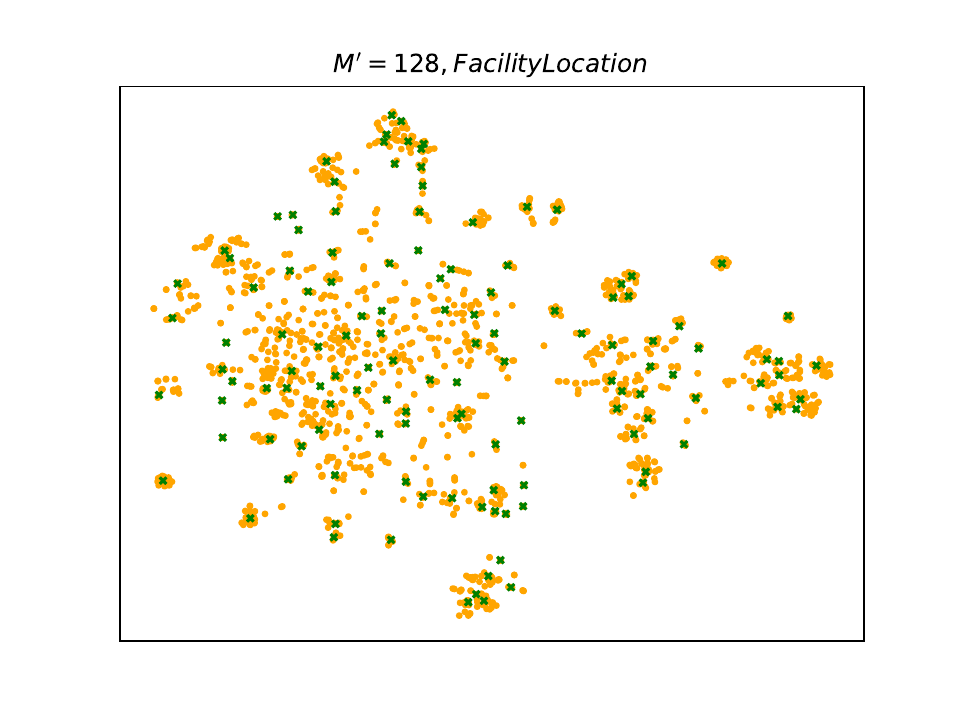}
    \end{subfigure}
    \begin{subfigure}[b]{0.46\linewidth}
        \centering
        \includegraphics[width=\linewidth]{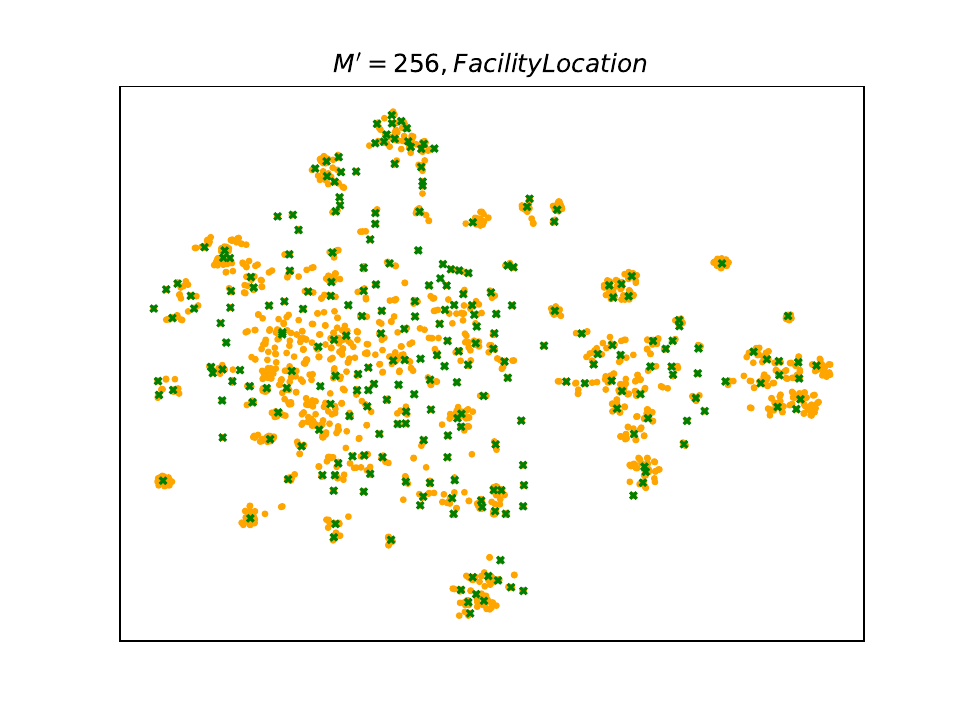}
    \end{subfigure}
    \begin{subfigure}[b]{0.46\linewidth}
        \centering
        \includegraphics[width=\linewidth]{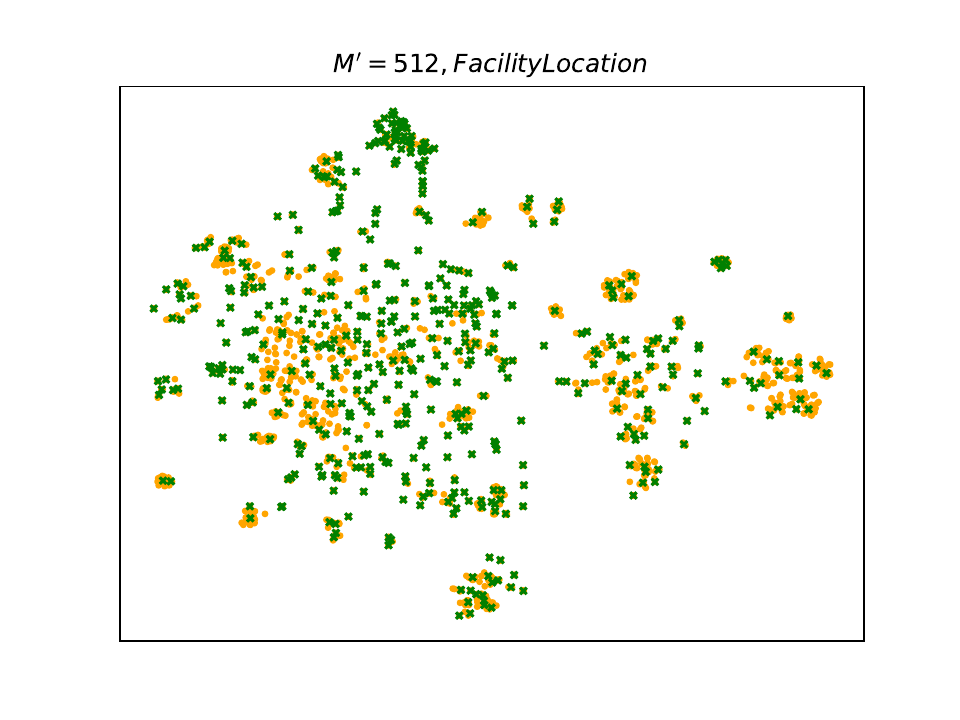}
    \end{subfigure}
    \begin{subfigure}[b]{0.46\linewidth}
        \centering
        \includegraphics[width=\linewidth]{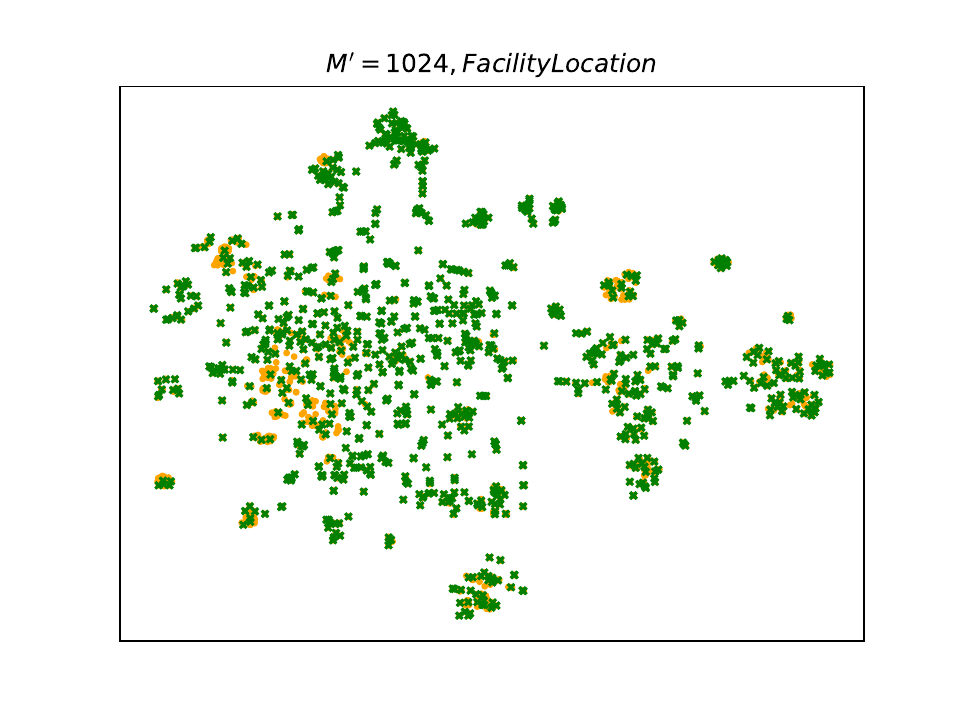}
    \end{subfigure}
    
    \caption{t-SNE visualizations for task subsets of various sizes ($M'$) selected by Facility Location}
    \label{fig:fl_visualization}
\end{figure*}

\begin{figure*}[h]
    \centering
    \begin{subfigure}[b]{0.46\linewidth}
        \centering
        \includegraphics[width=\linewidth]{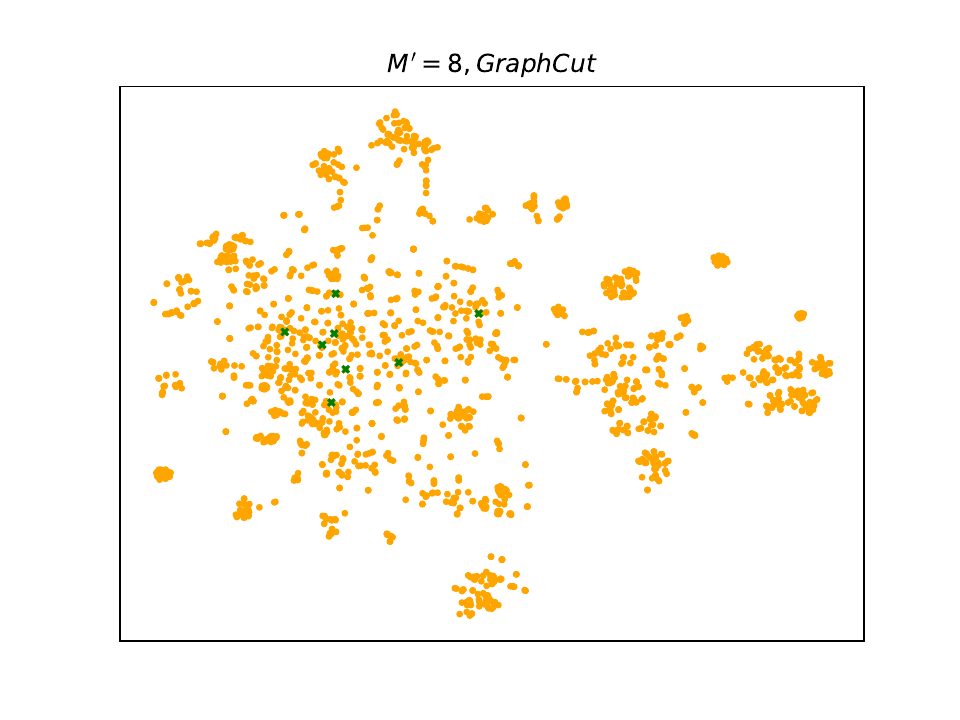}
    \end{subfigure}
    \begin{subfigure}[b]{0.46\linewidth}
        \centering
        \includegraphics[width=\linewidth]{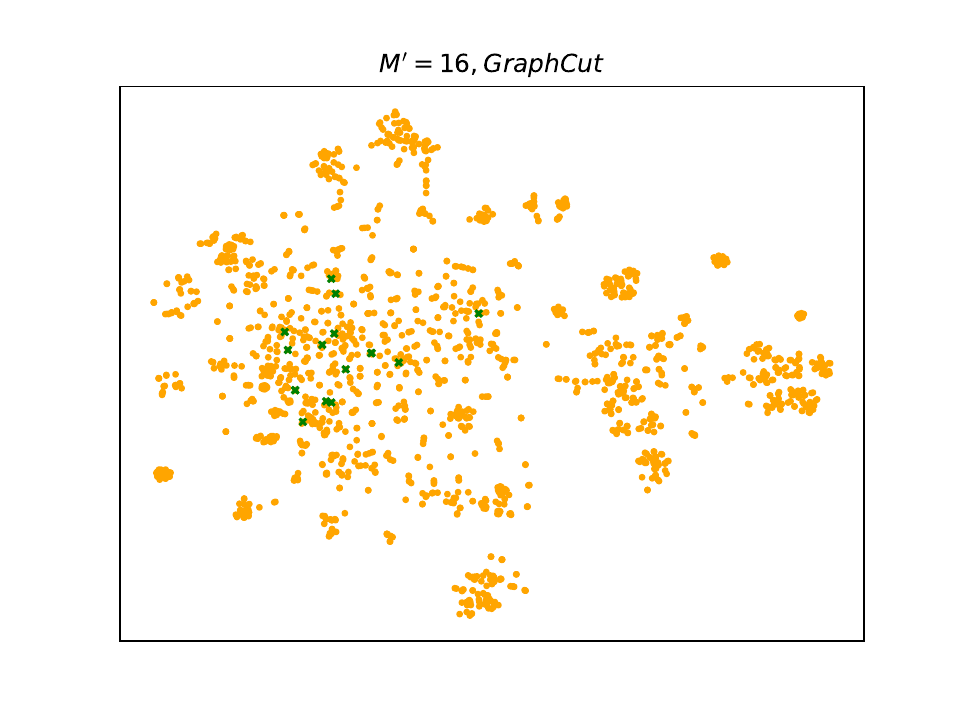}
    \end{subfigure}
    \begin{subfigure}[b]{0.46\linewidth}
        \centering
        \includegraphics[width=\linewidth]{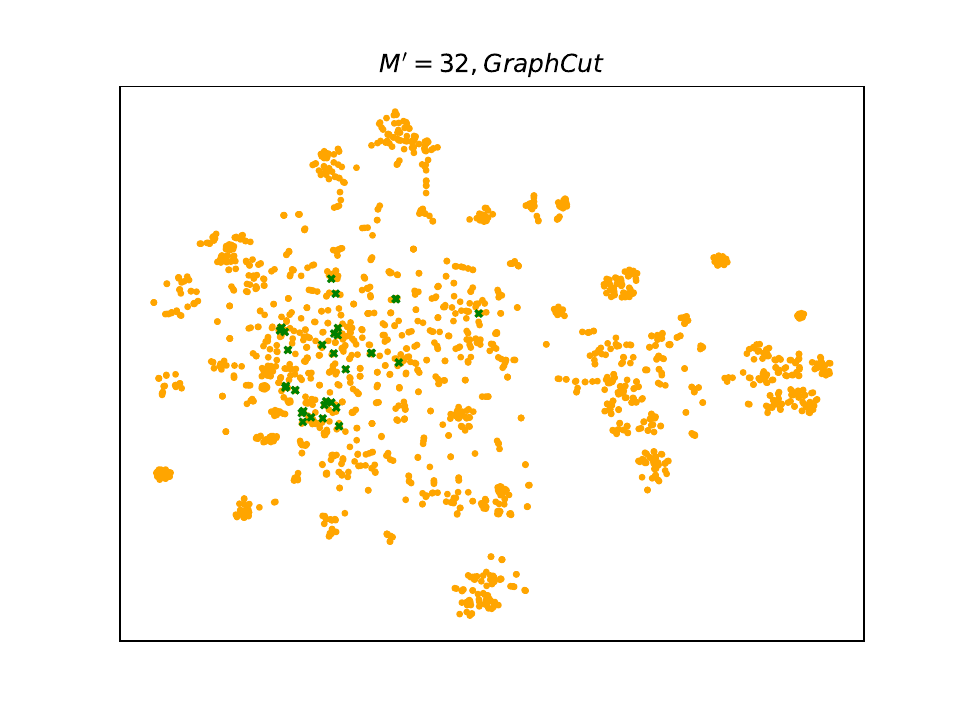}
    \end{subfigure}
    \begin{subfigure}[b]{0.46\linewidth}
        \centering
        \includegraphics[width=\linewidth]{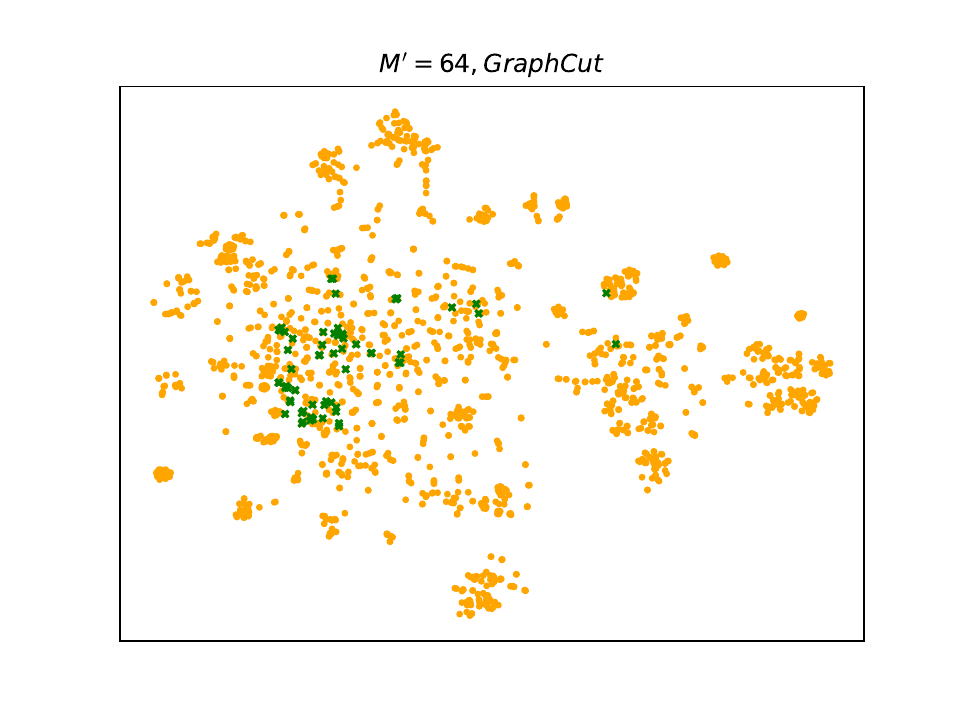}
    \end{subfigure}
    \begin{subfigure}[b]{0.46\linewidth}
        \centering
        \includegraphics[width=\linewidth]{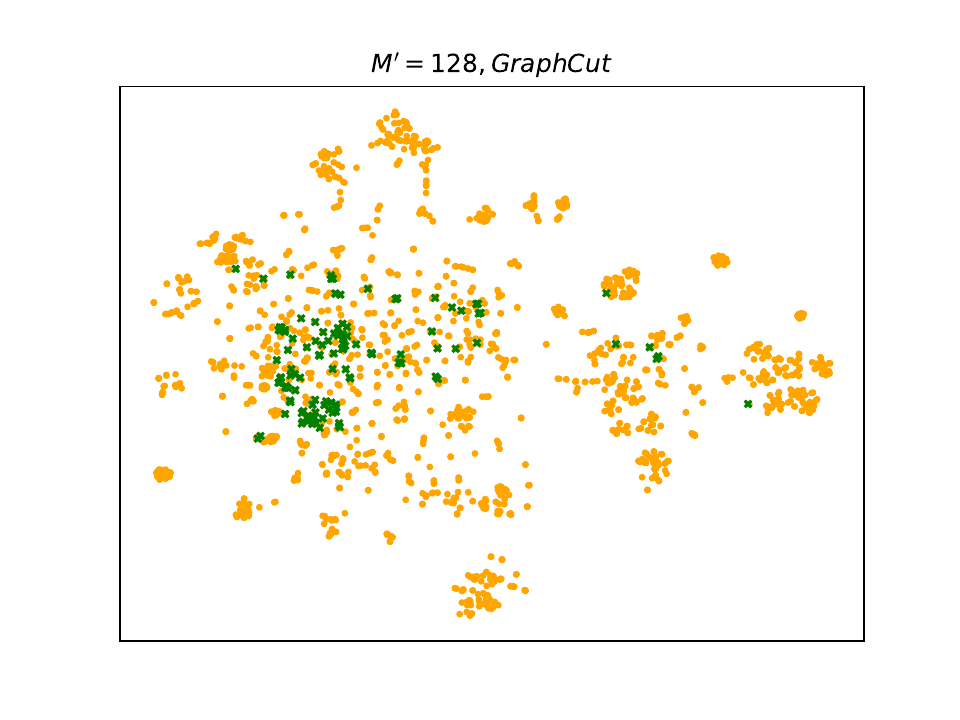}
    \end{subfigure}
    \begin{subfigure}[b]{0.46\linewidth}
        \centering
        \includegraphics[width=\linewidth]{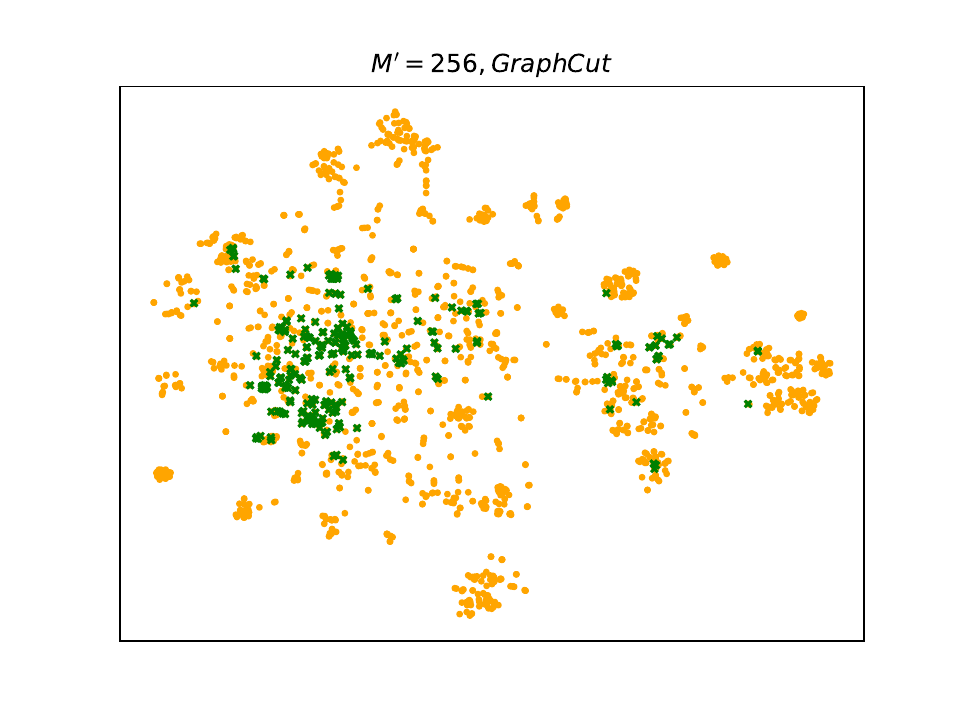}
    \end{subfigure}
    \begin{subfigure}[b]{0.46\linewidth}
        \centering
        \includegraphics[width=\linewidth]{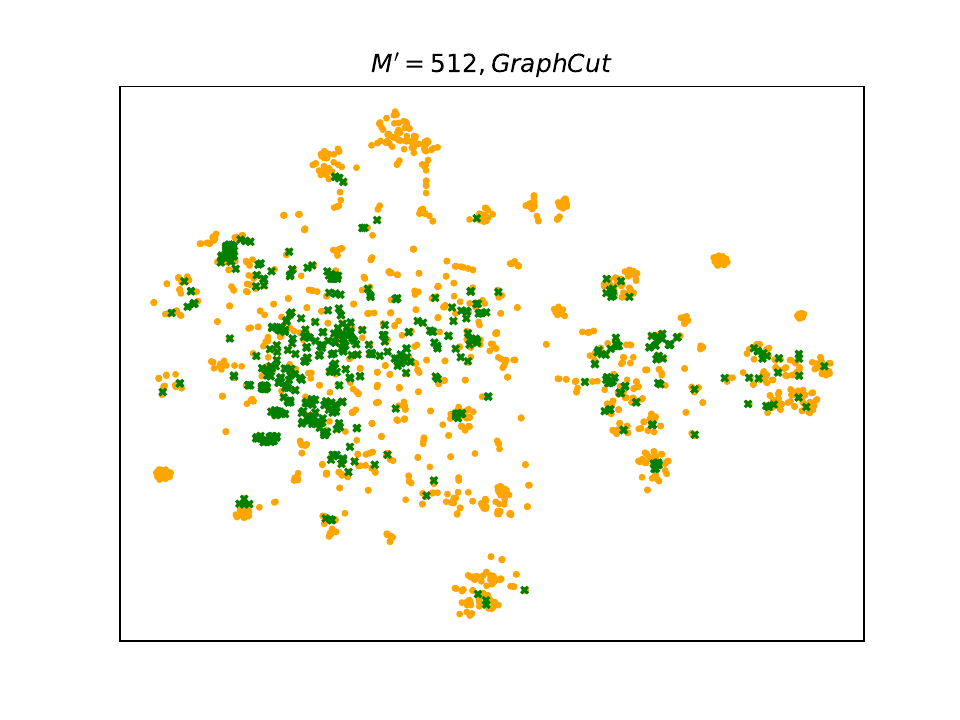}
    \end{subfigure}
    \begin{subfigure}[b]{0.46\linewidth}
        \centering
        \includegraphics[width=\linewidth]{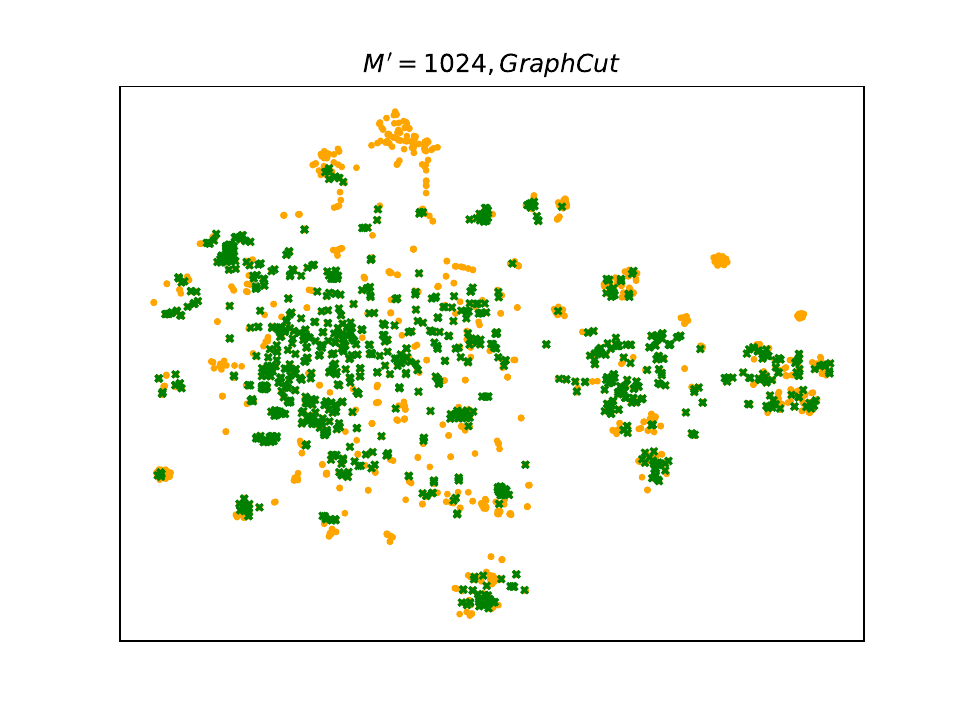}
    \end{subfigure}
    
    \caption{t-SNE visualizations for task subsets of various sizes ($M'$) selected by Graph Cut}
    \label{fig:gc_visualization}
\end{figure*}

\begin{figure*}[h]
    \centering
    \begin{subfigure}[b]{0.46\linewidth}
        \centering
        \includegraphics[width=\linewidth]{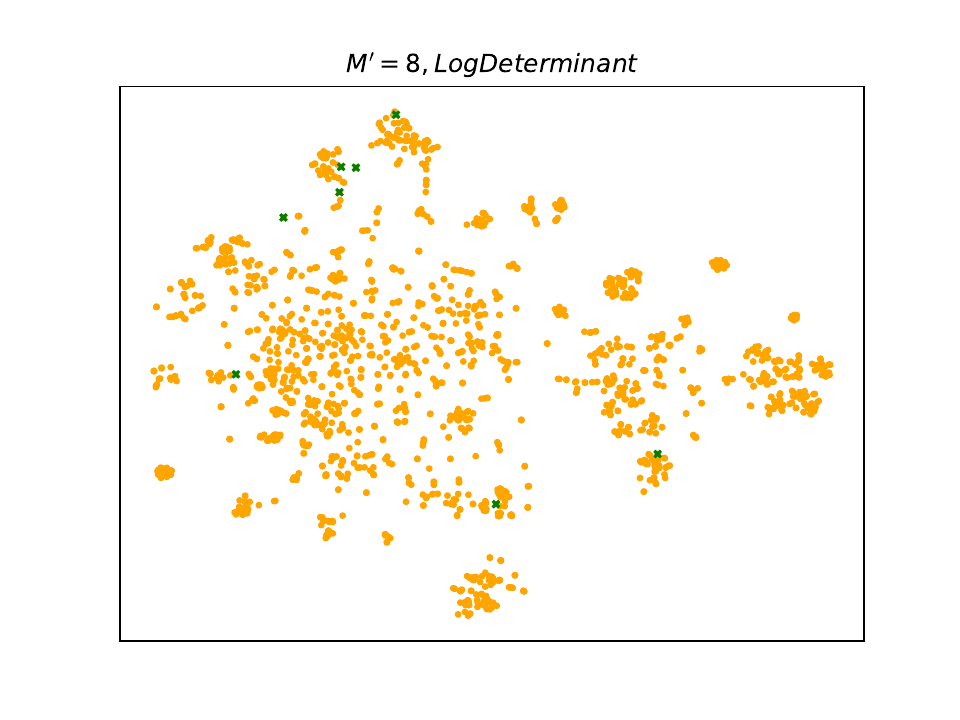}
    \end{subfigure}
    \begin{subfigure}[b]{0.46\linewidth}
        \centering
        \includegraphics[width=\linewidth]{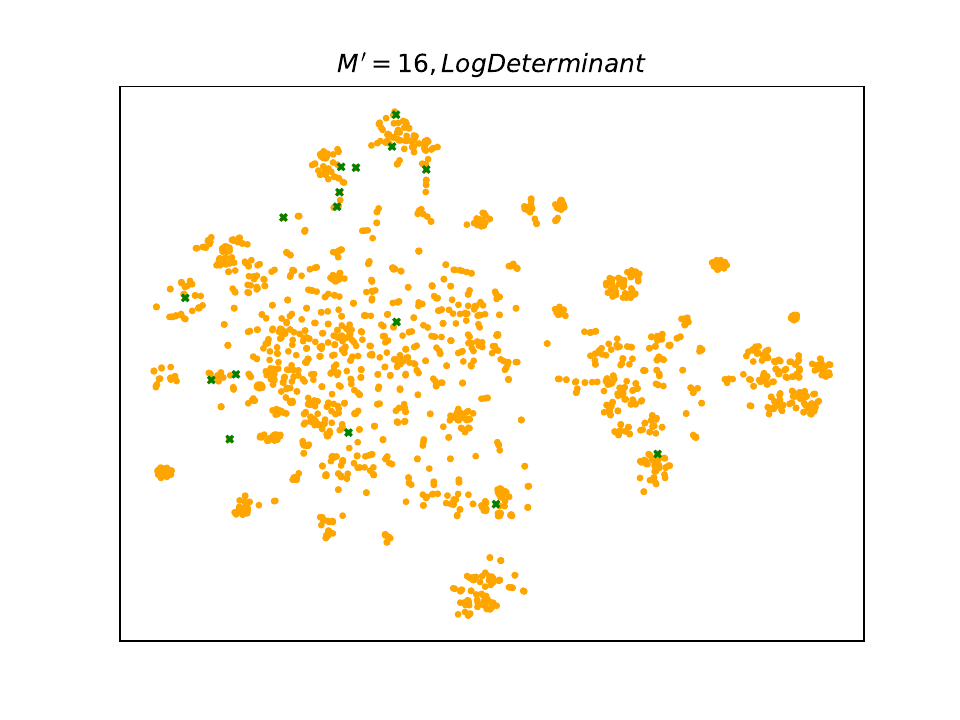}
    \end{subfigure}
    \begin{subfigure}[b]{0.46\linewidth}
        \centering
        \includegraphics[width=\linewidth]{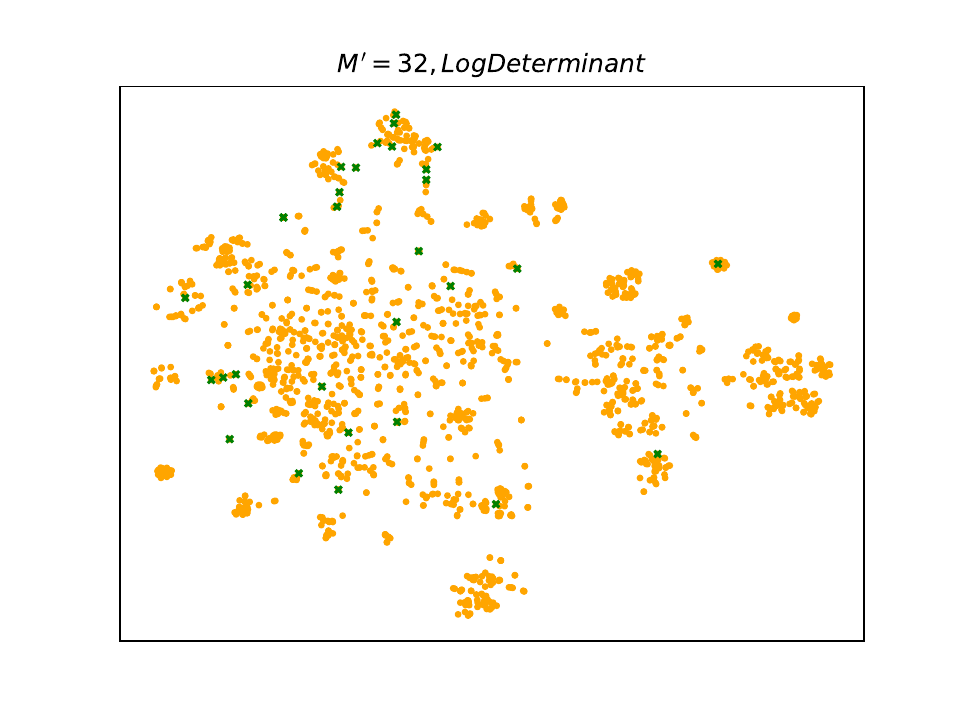}
    \end{subfigure}
    \begin{subfigure}[b]{0.46\linewidth}
        \centering
        \includegraphics[width=\linewidth]{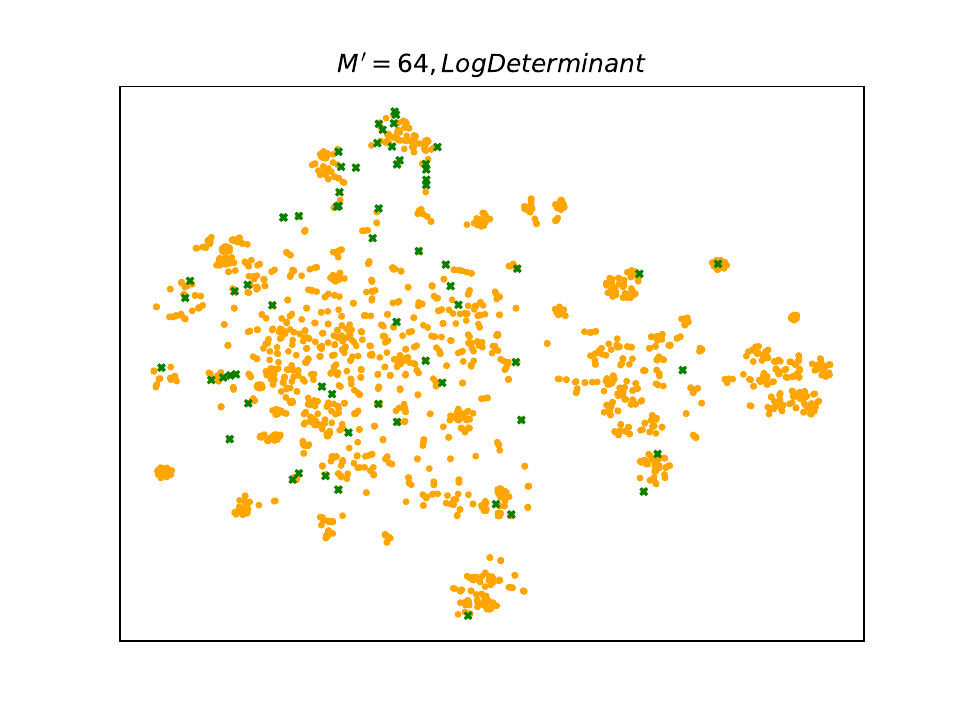}
    \end{subfigure}
    \begin{subfigure}[b]{0.46\linewidth}
        \centering
        \includegraphics[width=\linewidth]{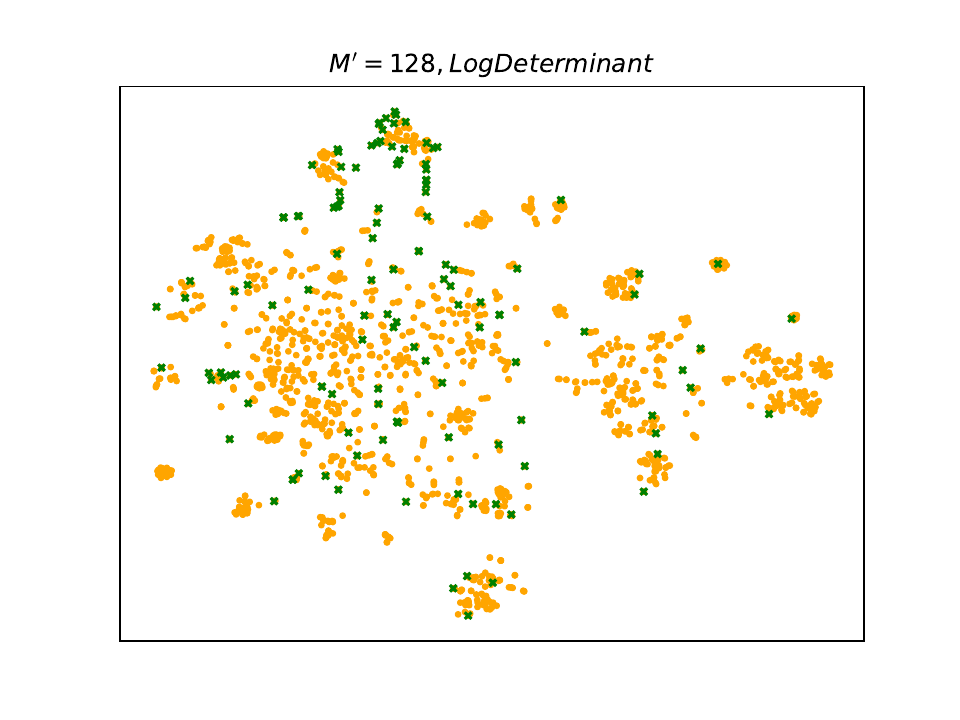}
    \end{subfigure}
    \begin{subfigure}[b]{0.46\linewidth}
        \centering
        \includegraphics[width=\linewidth]{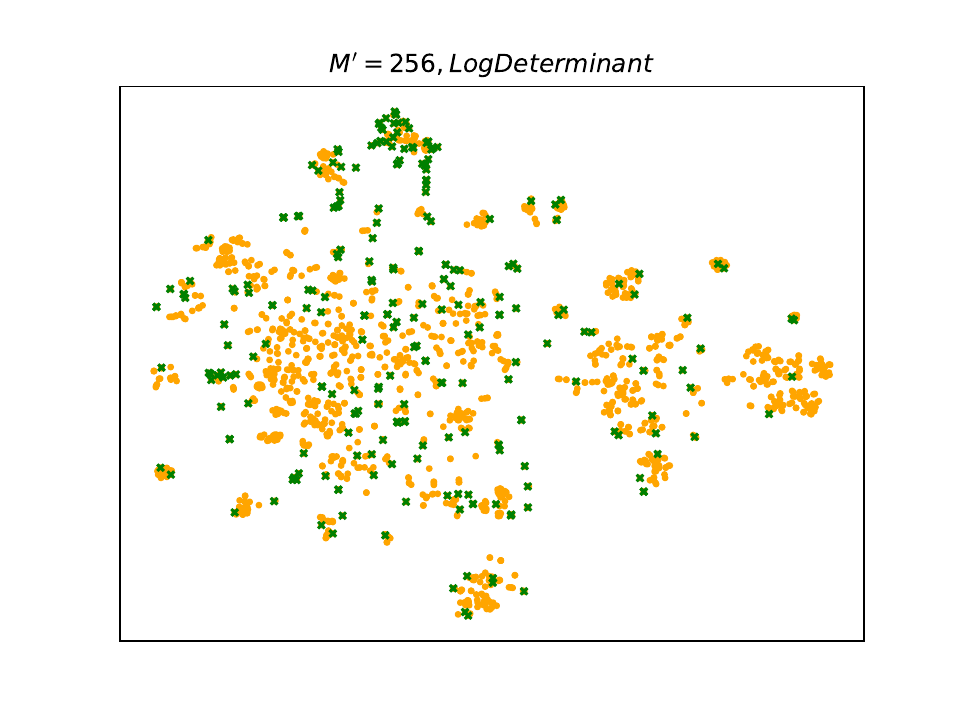}
    \end{subfigure}
    \begin{subfigure}[b]{0.46\linewidth}
        \centering
        \includegraphics[width=\linewidth]{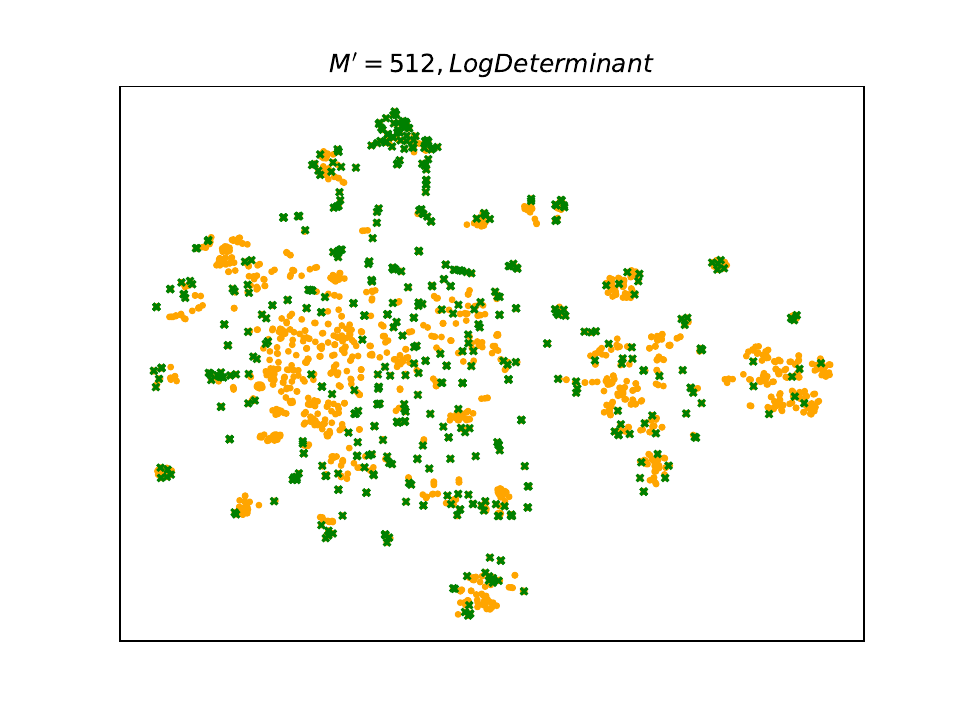}
    \end{subfigure}
    \begin{subfigure}[b]{0.46\linewidth}
        \centering
        \includegraphics[width=\linewidth]{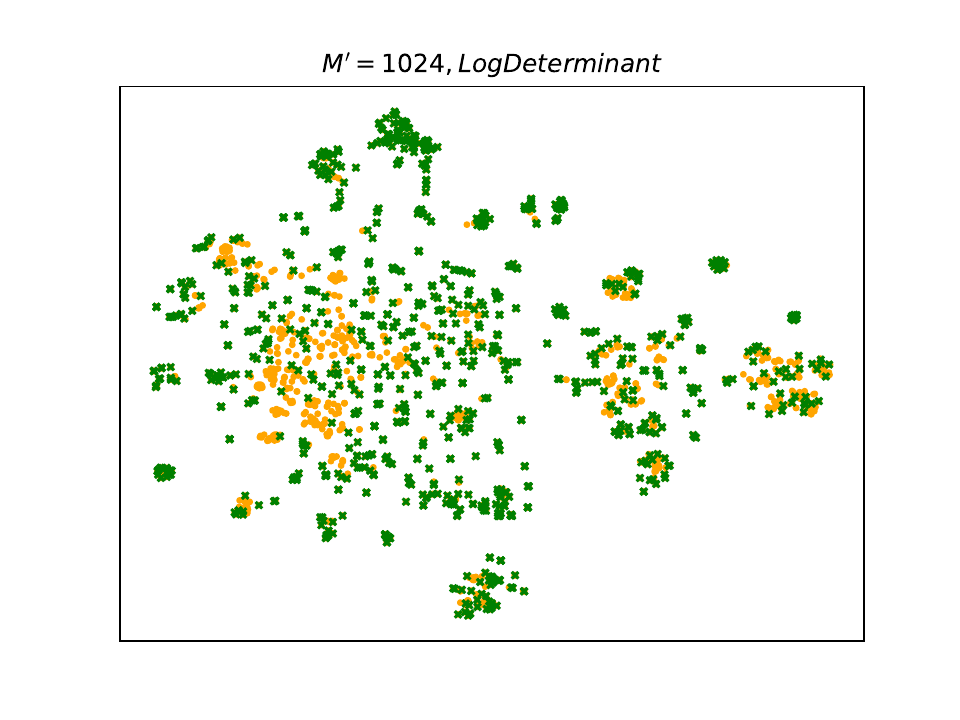}
    \end{subfigure}
    
    \caption{t-SNE visualizations for task subsets of various sizes ($M'$) selected by Log Determinant}
    \label{fig:logdet_visualization}
\end{figure*}

\begin{table*}
    \centering
\resizebox{\textwidth}{!}{
\begin{tabular}{cc}
\begin{tabular}{cl}
Rank & Task  \\
1 & anli/r3:0.1.0 \\ 
2 & hellaswag:1.1.0 \\ 
3 & task1392\_superglue\_multirc\_answer\_verification \\ 
4 & task955\_wiki\_auto\_style\_transfer \\ 
5 & task381\_boolq\_question\_generation \\ 
6 & task1291\_multi\_news\_summarization \\ 
7 & task1295\_adversarial\_qa\_question\_answering \\ 
8 & task519\_aquamuse\_question\_generation \\ 
9 & anli/r2:0.1.0 \\ 
10 & anli/r1:0.1.0 \\ 
11 & race\_high\_Select\_the\_best\_answer\_no\_instructions\_ \\ 
12 & cot\_ecqa\_ii \\ 
13 & super\_glue/rte:1.0.2 \\ 
14 & task1660\_super\_glue\_question\_generation \\ 
15 & task339\_record\_answer\_generation \\ 
16 & task302\_record\_classification \\ 
17 & stream\_qed\_ii \\ 
18 & task870\_msmarco\_answer\_generation \\ 
19 & paws\_wiki:1.1.0 \\ 
20 & super\_glue/multirc:1.0.2 \\ 
21 & task603\_wikitext-103\_fill\_in\_the\_blank \\ 
22 & task887\_quail\_answer\_generation \\ 
23 & stream\_qed \\ 
24 & task380\_boolq\_yes\_no\_question \\ 
25 & coqa:1.0.0 \\ 
26 & task1412\_web\_questions\_question\_answering \\ 
27 & super\_glue/cb:1.0.2 \\ 
28 & task1293\_kilt\_tasks\_hotpotqa\_question\_answering \\ 
29 & quail\_context\_question\_description\_answer\_text \\ 
30 & fix\_punct \\ 
31 & true\_case \\ 
32 & winogrande:1.1.0 \\ 
33 & glue/wnli:2.0.0 \\ 
34 & super\_glue/record:1.0.2 \\ 
35 & cot\_ecqa \\ 
36 & quail\_context\_question\_description\_answer\_id \\ 
37 & task919\_coqa\_incorrect\_answer\_generation \\ 
38 & task520\_aquamuse\_answer\_given\_in\_passage \\ 
39 & task1290\_xsum\_summarization \\ 
40 & task1609\_xquad\_en\_question\_generation \\ 
41 & squad/v1.1:3.0.0 \\ 
42 & task231\_iirc\_link\_classification \\ 
43 & task349\_squad2.0\_answerable\_unanswerable\_question\_classification \\ 
44 & wiki\_dialog\_ii \\ 
45 & task1661\_super\_glue\_classification \\ 
46 & quail\_context\_description\_question\_text \\ 
47 & adversarial\_qa\_droberta\_answer\_the\_following\_q \\ 
48 & task644\_refresd\_translation \\ 
49 & bool\_q:1.0.0 \\ 
50 & task470\_mrqa\_question\_generation \\ 
51 & race\_middle\_Select\_the\_best\_answer\_no\_instructions\_ \\ 
52 & task770\_pawsx\_english\_text\_modification \\ 
53 & adversarial\_qa\_dbidaf\_answer\_the\_following\_q \\ 
54 & glue/qnli:2.0.0 \\ 
55 & task1558\_jfleg\_incorrect\_answer\_generation \\ 
56 & task344\_hybridqa\_answer\_generation \\ 
57 & quoref\_Guess\_Title\_For\_Context \\ 
58 & super\_glue/copa:1.0.2 \\ 
59 & word\_segment \\ 
60 & task1340\_msr\_text\_compression\_compression \\ 
61 & wiki\_dialog \\ 
62 & squad/v2.0:3.0.0 \\ 
63 & task1530\_scitail1.1\_sentence\_generation \\ 
64 & task225\_english\_language\_answer\_generation \\ 
\end{tabular}
& 
\begin{tabular}{cl}
Rank & Task \\
65 & task303\_record\_incorrect\_answer\_generation \\ 
66 & gem/web\_nlg\_en:1.1.0 \\ 
67 & task768\_qed\_text\_span\_selection \\ 
68 & adversarial\_qa\_dbert\_answer\_the\_following\_q \\ 
69 & task1389\_hellaswag\_completion \\ 
70 & task1294\_wiki\_qa\_answer\_verification \\ 
71 & wiki\_qa\_Topic\_Prediction\_Answer\_Only \\ 
72 & task596\_mocha\_question\_generation \\ 
73 & task871\_msmarco\_question\_generation \\ 
74 & task1564\_triviaqa\_answer\_generation \\ 
75 & task1344\_glue\_entailment\_classification \\ 
76 & cosmos\_qa:1.0.0 \\ 
77 & task1555\_scitail\_answer\_generation \\ 
78 & task1557\_jfleg\_answer\_generation \\ 
79 & gem/common\_gen:1.1.0 \\ 
80 & task238\_iirc\_answer\_from\_passage\_answer\_generation \\ 
81 & task233\_iirc\_link\_exists\_classification \\ 
82 & super\_glue/wic:1.0.2 \\ 
83 & task1345\_glue\_qqp\_question\_paraprashing \\ 
84 & glue/qqp:2.0.0 \\ 
85 & glue/stsb:2.0.0 \\ 
86 & task595\_mocha\_answer\_generation \\ 
87 & task460\_qasper\_answer\_generation \\ 
88 & glue/mnli:2.0.0 \\ 
89 & task051\_multirc\_correct\_answer\_single\_sentence \\ 
90 & adversarial\_qa\_droberta\_tell\_what\_it\_is \\ 
91 & task547\_alt\_translation\_entk\_en \\ 
92 & quail\_no\_prompt\_id \\ 
93 & task311\_race\_question\_generation \\ 
94 & cot\_sensemaking\_ii \\ 
95 & gem/dart:1.1.0 \\ 
96 & wiki\_qa\_Jeopardy\_style \\ 
97 & adversarial\_qa\_dbidaf\_tell\_what\_it\_is \\ 
98 & task1593\_yahoo\_answers\_topics\_classification \\ 
99 & quail\_context\_question\_answer\_description\_text \\ 
100 & cot\_creak\_ii \\ 
101 & definite\_pronoun\_resolution:1.1.0 \\ 
102 & gigaword:1.2.0 \\ 
103 & super\_glue/wsc.fixed:1.0.2 \\ 
104 & task234\_iirc\_passage\_line\_answer\_generation \\ 
105 & task556\_alt\_translation\_en\_ja \\ 
106 & task604\_flores\_translation\_entosn \\ 
107 & adversarial\_qa\_dbert\_tell\_what\_it\_is \\ 
108 & task310\_race\_classification \\ 
109 & task1594\_yahoo\_answers\_topics\_question\_generation \\ 
110 & story\_cloze/2016:1.0.0 \\ 
111 & task933\_wiki\_auto\_style\_transfer \\ 
112 & gem/wiki\_lingua\_english\_en:1.1.0 \\ 
113 & task1382\_quarel\_write\_correct\_answer \\ 
114 & task1296\_wiki\_hop\_question\_answering \\ 
115 & quail\_context\_description\_question\_answer\_text \\ 
116 & glue/cola:2.0.0 \\ 
117 & cot\_strategyqa\_ii \\ 
118 & task1553\_cnn\_dailymail\_summarization \\ 
119 & openbookqa:0.1.0 \\ 
120 & task054\_multirc\_write\_correct\_answer \\ 
121 & task1218\_ted\_translation\_en\_ja \\ 
122 & quail\_no\_prompt\_text \\ 
123 & quail\_context\_question\_description\_text \\ 
124 & task550\_discofuse\_sentence\_generation \\ 
125 & quail\_context\_question\_answer\_description\_id \\ 
126 & task1608\_xquad\_en\_answer\_generation \\ 
127 & task1520\_qa\_srl\_answer\_generation \\ 
128 & cos\_e\_v1.11\_generate\_explanation\_given\_text \\ 
\end{tabular}
\end{tabular}
}
\caption{List of 128 most representative tasks in FLAN-2022 collection as ordered by the Graph Cut}
\label{tab:gains_table}
\end{table*}

\begin{table*}
    \centering
\resizebox{\textwidth}{!}{
\begin{tabular}{cc}
\begin{tabular}{cl}
Rank & Task  \\
1713 & task261\_spl\_translation\_es\_en \\ 
1714 & task1384\_deal\_or\_no\_dialog\_classification \\ 
1715 & task854\_hippocorpus\_classification \\ 
1716 & task110\_logic2text\_sentence\_generation \\ 
1717 & task360\_spolin\_yesand\_response\_generation \\ 
1718 & task148\_afs\_argument\_quality\_gay\_marriage \\ 
1719 & task499\_extract\_and\_add\_all\_numbers\_from\_list \\ 
1720 & task176\_break\_decompose\_questions \\ 
1721 & task085\_unnatural\_addsub\_arithmetic \\ 
1722 & task108\_contextualabusedetection\_classification \\ 
1723 & task472\_haspart\_classification \\ 
1724 & task856\_conv\_ai\_2\_classification \\ 
1725 & task600\_find\_the\_longest\_common\_substring\_in\_two\_strings \\ 
1726 & task150\_afs\_argument\_quality\_gun\_control \\ 
1727 & task1508\_wordnet\_antonyms \\ 
1728 & task183\_rhyme\_generation \\ 
1729 & task488\_extract\_all\_alphabetical\_elements\_from\_list\_in\_order \\ 
1730 & task682\_online\_privacy\_policy\_text\_classification \\ 
1731 & task1425\_country\_iso\_numeric \\ 
1732 & task756\_find\_longert\_substring\_and\_return\_all\_unique\_alphabets\_in\_it \\ 
1733 & task1585\_root09\_hypernym\_generation \\ 
1734 & task958\_e2e\_nlg\_text\_generation\_parse \\ 
1735 & task584\_udeps\_eng\_fine\_pos\_tagging \\ 
1736 & task1319\_country\_by\_barcode\_prefix \\ 
1737 & task1507\_boolean\_temporal\_reasoning \\ 
1738 & task509\_collate\_of\_all\_alphabetical\_and\_numerical\_elements\_in\_list\_separately \\ 
1739 & task064\_all\_elements\_except\_first\_i \\ 
1740 & task130\_scan\_structured\_text\_generation\_command\_action\_long \\ 
1741 & task365\_synthetic\_remove\_vowels \\ 
1742 & task149\_afs\_argument\_quality\_death\_penalty \\ 
1743 & task210\_logic2text\_structured\_text\_generation \\ 
1744 & task1495\_adverse\_drug\_event\_classification \\ 
1745 & task684\_online\_privacy\_policy\_text\_information\_type\_generation \\ 
1746 & task1426\_country\_independence\_year \\ 
1747 & task126\_scan\_structured\_text\_generation\_command\_action\_all \\ 
1748 & task605\_find\_the\_longest\_common\_subsequence\_in\_two\_lists \\ 
1749 & task128\_scan\_structured\_text\_generation\_command\_action\_short \\ 
1750 & task960\_ancora-ca-ner\_named\_entity\_recognition \\ 
1751 & task078\_all\_elements\_except\_last\_i \\ 
1752 & task1427\_country\_region\_in\_world \\ 
1753 & task063\_first\_i\_elements \\ 
1754 & task956\_leetcode\_420\_strong\_password\_check \\ 
1755 & task683\_online\_privacy\_policy\_text\_purpose\_answer\_generation \\ 
1756 & task091\_all\_elements\_from\_index\_i\_to\_j \\ 
1757 & task1542\_every\_ith\_element\_from\_starting \\ 
1758 & task1506\_celebrity\_minimal\_dob\_span \\ 
1759 & task245\_check\_presence\_in\_set\_intersection \\ 
1760 & task497\_extract\_all\_numbers\_from\_list\_in\_order \\ 
1761 & task1428\_country\_surface\_area \\ 
1762 & task092\_check\_prime\_classification \\ 
1763 & task1088\_array\_of\_products \\ 
1764 & task1332\_check\_leap\_year \\ 
1765 & task127\_scan\_long\_text\_generation\_action\_command\_all \\ 
1766 & task129\_scan\_long\_text\_generation\_action\_command\_short \\ 
1767 & task1322\_country\_government\_type \\ 
1768 & task1331\_reverse\_array \\ 
1769 & task131\_scan\_long\_text\_generation\_action\_command\_long \\ 
1770 & task371\_synthetic\_product\_of\_list \\ 
1771 & task1189\_check\_char\_in\_string \\ 
1772 & task208\_combinations\_of\_list \\ 
1773 & task211\_logic2text\_classification \\ 
1774 & task1551\_every\_ith\_element\_from\_kth\_element \\ 
1775 & task1194\_kth\_largest\_element \\ 
1776 & task101\_reverse\_and\_concatenate\_all\_elements\_from\_index\_i\_to\_j \\ 
\end{tabular}
& 
\begin{tabular}{cl}
Rank & Task \\
1777 & task1404\_date\_conversion \\ 
1778 & task504\_count\_all\_alphabetical\_elements\_in\_list \\ 
1779 & task087\_new\_operator\_addsub\_arithmetic \\ 
1780 & task850\_synthetic\_longest\_palindrome \\ 
1781 & task099\_reverse\_elements\_between\_index\_i\_and\_j \\ 
1782 & task206\_collatz\_conjecture \\ 
1783 & task505\_count\_all\_numerical\_elements\_in\_list \\ 
1784 & task1405\_find\_median \\ 
1785 & task267\_concatenate\_and\_reverse\_all\_elements\_from\_index\_i\_to\_j \\ 
1786 & task207\_max\_element\_lists \\ 
1787 & task1443\_string\_to\_number \\ 
1788 & task1188\_count\_max\_freq\_char \\ 
1789 & task212\_logic2text\_classification \\ 
1790 & task374\_synthetic\_pos\_or\_neg\_calculation \\ 
1791 & task1190\_add\_integer\_to\_list \\ 
1792 & task243\_count\_elements\_in\_set\_intersection \\ 
1793 & task636\_extract\_and\_sort\_unique\_alphabets\_in\_a\_list \\ 
1794 & task124\_conala\_pair\_averages \\ 
1795 & task1150\_delete\_max\_min \\ 
1796 & task755\_find\_longest\_substring\_and\_replace\_its\_sorted\_lowercase\_version\_in\_both\_lists \\ 
1797 & task100\_concatenate\_all\_elements\_from\_index\_i\_to\_j \\ 
1798 & task372\_synthetic\_palindrome\_numbers \\ 
1799 & task1148\_maximum\_ascii\_value \\ 
1800 & task506\_position\_of\_all\_alphabetical\_elements\_in\_list \\ 
1801 & task373\_synthetic\_round\_tens\_place \\ 
1802 & task367\_synthetic\_remove\_floats \\ 
1803 & task1406\_kth\_smallest\_element \\ 
1804 & task1333\_check\_validity\_date\_ddmmyyyy \\ 
1805 & task244\_count\_elements\_in\_set\_union \\ 
1806 & task205\_remove\_even\_elements \\ 
1807 & task1320\_country\_domain\_tld \\ 
1808 & task123\_conala\_sort\_dictionary \\ 
1809 & task122\_conala\_list\_index\_addition \\ 
1810 & task076\_splash\_correcting\_sql\_mistake \\ 
1811 & task094\_conala\_calculate\_mean \\ 
1812 & task507\_position\_of\_all\_numerical\_elements\_in\_list \\ 
1813 & task1403\_check\_validity\_date\_mmddyyyy \\ 
1814 & task1315\_find\_range\_array \\ 
1815 & task098\_conala\_list\_intersection \\ 
1816 & task1087\_two\_number\_sum \\ 
1817 & task095\_conala\_max\_absolute\_value \\ 
1818 & task1089\_check\_monotonic\_array \\ 
1819 & task077\_splash\_explanation\_to\_sql \\ 
1820 & task097\_conala\_remove\_duplicates \\ 
1821 & task125\_conala\_pair\_differences \\ 
1822 & task368\_synthetic\_even\_or\_odd\_calculation \\ 
1823 & task1151\_swap\_max\_min \\ 
1824 & task852\_synthetic\_multiply\_odds \\ 
1825 & task606\_sum\_of\_all\_numbers\_in\_list\_between\_positions\_i\_and\_j \\ 
1826 & task090\_equation\_learner\_algebra \\ 
1827 & task1446\_farthest\_integers \\ 
1828 & task096\_conala\_list\_index\_subtraction \\ 
1829 & task868\_cfq\_mcd1\_explanation\_to\_sql \\ 
1830 & task369\_synthetic\_remove\_odds \\ 
1831 & task093\_conala\_normalize\_lists \\ 
1832 & task370\_synthetic\_remove\_divisible\_by\_3 \\ 
1833 & task851\_synthetic\_multiply\_evens \\ 
1834 & task637\_extract\_and\_sort\_unique\_digits\_in\_a\_list \\ 
1835 & task1498\_24hour\_to\_12hour\_clock \\ 
1836 & task869\_cfq\_mcd1\_sql\_to\_explanation \\ 
1837 & task1445\_closest\_integers \\ 
1838 & task1444\_round\_power\_of\_two \\ 
1839 & task366\_synthetic\_return\_primes \\ 
1840 & task107\_splash\_question\_to\_sql \\ 
\end{tabular}
\end{tabular}
}
\caption{List of 128 least representative tasks in FLAN-2022 collection as ordered by the Graph Cut}
\label{tab:gains_table_last_128}
\end{table*}

As pointed out in Section~\ref{sec:background_submodularity}, the three submodular functions in Table~\ref{tab:submodular_functions_examples} are different. Facility Location predominantly models representation; Graph Cut models a trade-off between representation and diversity; Log Determinant predominantly models diversity. To better visualize this, we present t-SNE plots of 1840 tasks present in the FLAN 2022 collection~\citep{longpre2023flan} and highlight the tasks selected by these functions for different values of $M'$. Figure~\ref{fig:fl_visualization}, Figure~\ref{fig:gc_visualization} and Figure~\ref{fig:logdet_visualization} respectively contains the visualizations for Facility Location, Graph Cut and Log Determinant.

\end{document}